\documentclass{article}

\usepackage{PRIMEarxiv}

\usepackage[utf8]{inputenc} 
\usepackage[T1]{fontenc}    
\usepackage{hyperref}       
\usepackage{url}            
\usepackage{booktabs}       
\usepackage{amsfonts}       
\usepackage{amsmath}
\usepackage{amssymb}
\usepackage{mathtools} 
\usepackage{nicefrac}       
\usepackage{microtype}      
\usepackage{fancyhdr}       
\usepackage{graphicx}       
\usepackage{multirow}
\usepackage{graphicx}
\usepackage{array}
\usepackage{float}
\usepackage{placeins}
\usepackage{algorithm}
\usepackage{algpseudocode}
\graphicspath{{media/}}     

\title{Exploring the Potential of Probabilistic Transformer\\ for Time Series Modeling:\\ A Report on the ST-PT Framework
}

\author{
  Zhangzhi Xiong, Haoyi Wu, You Wu, Shuqi Gu, 
  Kan Ren\thanks{Corresponding authors.}, 
  Kewei Tu\footnotemark[1] \\
  School of Information Science and Technology, ShanghaiTech University \\
  Shanghai Engineering Research Center of Intelligent Vision and Imaging\\
  \texttt{\{xiongzhzh2023, wuhy1, wuyou2024, gushq2024, renkan, tukw\}@shanghaitech.edu.cn} \\
}

\begin{document}
\maketitle

\begin{abstract}
The \textit{\textbf{P}robabilistic \textbf{T}ransformer} (PT) establishes that the Transformer's self-attention plus its feed-forward block is mathematically equivalent to \textit{\textbf{M}ean-\textbf{F}ield \textbf{V}ariational \textbf{I}nference} (MFVI) on a \textit{\textbf{C}onditional \textbf{R}andom \textbf{F}ield} (CRF). Under this equivalence the Transformer ceases to be a black-box neural network and becomes a programmable factor graph: graph topology, factor potentials, and the message-passing schedule are all explicit and inspectable primitives that can be engineered. PT was originally developed for natural language and in this report we investigate its potential for time series. We first lift PT into the \emph{\textbf{S}patio-\textbf{T}emporal \textbf{P}robabilistic \textbf{T}ransformer} (ST-PT)---a two-dimensional factor graph over channels and time patches---to repair PT's missing channel axis and weak per-step semantics, and adopt ST-PT as a shared cornerstone backbone. We then identify three distinct properties that PT/ST-PT offers as a factor-graph model and derive three Research Questions, one per property, that probe how each property can be exploited in time series:
\textbf{RQ1.} The graph topology and potentials are direct programmable primitives. Can this be used to inject symbolic time-series priors (e.g. periodicity, trend) into ST-PT through structural graph modifications, especially under data scarcity and noise?
\textbf{RQ2.} The CRF's factor matrices \emph{are} the operator's potentials. Can an external condition program these factor matrices on a per-sample basis, so that conditional generation becomes structural---instantiating a sample-specific operator---rather than feature-level modulation of a fixed one?
\textbf{RQ3.} Each MFVI iteration is a Bayesian posterior update on the factor graph. Can this turn the latent transition of latent-space \textit{\textbf{A}uto\textbf{R}egressive} (AR) forecasting from an opaque MLP into a principled posterior update, and can a CRF teacher distill its latents into the AR student to counter cumulative error?
We give one empirical study per question---few-shot synthetic and real-data forecasting (RQ1), conditional time series generation with our PT-FG(\textit{\textbf{F}actor \textbf{G}eneration}) model on ConTSG-Bench (RQ2), and long-horizon latent-space AR forecasting against classical AR baselines (RQ3). Together, the three studies position ST-PT as a programmable framework whose three levers---graph topology, factor parameterization, and inference protocol---each provide a distinct and decent handle on a different aspect of time series modelling.
\end{abstract}

\section{Introduction}

\subsection{Background: An Architecture-Centric Paradigm}

Most contemporary deep time series modeling operates under what we will call the \emph{architecture-centric paradigm}. The architecture is, in effect, the only knob through which the model can be programmed, and every new piece of structural knowledge demands a new structural trick. A direct consequence is that capability and interface are conflated: one cannot inject a prior, supply a condition, or reshape the inference protocol without inventing a new architecture and hoping the model learns to use it correctly from data alone. 

The landscape of \textit{\textbf{M}ultivariate \textbf{T}ime \textbf{S}eries \textbf{F}orecasting} (MTSF) has shifted decisively from classical statistical models to deep neural architectures~\cite{ts_dl}. Traditional tools such as STL decomposition~\cite{stl} and ARIMA~\cite{arima} provided interpretability and explicit mechanisms for encoding structural beliefs, but struggled with nonlinearity and high dimensionality. Contemporary research is dominated by architectural innovation: decomposition-based Transformers~\cite{autoformer,fedformer}, efficient long-sequence attention~\cite{informer}, patch-based tokenization~\cite{patchtst}, and channel-as-token inversion~\cite{iTransformer}, among many others. Most of this work operates under a single implicit assumption: that a sufficiently expressive, purely data-driven architecture can learn all necessary temporal dynamics from sufficiently large datasets. This assumption has drawn scrutiny. DLinear~\cite{dlinear} notably showed that a one-layer linear baseline frequently outperforms sophisticated Transformers, and subsequent analyses~\cite{transformerfail} argue that attention's permutation invariance is fundamentally at odds with the sequential and physical nature of time series, yielding overfitting in noisy or small-sample regimes.

The same architecture-centric pattern recurs in time series tasks beyond MTSF. In conditional time series generation, contemporary methods inject the external condition through feature-level modulation of a fixed denoiser---adaptive LayerNorm, cross-attention, classifier-free guidance~\cite{narasimhan2024timeweaver,gu2025verbalts,ge2025t2s,li2025bridge}---so the network topology is left unchanged and the condition only shifts hidden activations of an otherwise generic backbone. In autoregressive forecasting, both output-space~\cite{salinas2019deepvar,hochreiter1997lstm,lai2018lstnet} and latent-space~\cite{nextlat} approaches drive the step-to-step transition through a free-form learned module with no explicit Bayesian semantics. Across all three task families, the field lacks a shared, explicit interface through which they can be inspected or engineered.

\subsection{The Probabilistic Transformer and Its Three Primitives}
\label{3primitive}
The \textit{\textbf{P}robabilistic \textbf{T}ransformer} (PT)~\cite{pt} establishes a pivotal equivalence: the Transformer's multi-head self-attention plus its feed-forward block is mathematically equivalant to the mean-field variational inference (MFVI) update of a dependency conditional random field (CRF). Under this re-reading, tokens become latent variables $Z$, attention heads become per-head dependency variables $H$, and the feed-forward block corresponds to a global topic factor with auxiliary variables $G$; unary, ternary, and binary factor matrices are the CRF's potentials. In this framework, the Transformer transcends its black-box origins, emerging instead as an undirected factor graph governed by a principled inference procedure. By treating graph topology, factor potentials, and the schedule of message-passing iterations as explicit primitives, we move beyond simple end-to-end learning toward a regime of deliberate architectural engineering.

This re-reading exposes three programmable primitives that are largely invisible from the standard Transformer view:
\begin{itemize}
\item \textbf{(P1) Graph topology and potentials are explicit primitives.} A factor graph is built from explicit nodes, edges, and unary / pairwise / higher-order potentials. One can add a latent node, modulate a pairwise potential, or insert a new cross-variable edge without redesigning the architecture.
\item \textbf{(P2) Factor matrices are the operator's potentials.} In an ordinary Transformer, weight matrices are interior parameters of an opaque function. Under the CRF view, the corresponding factor matrices \emph{define} the message-passing operator itself, and sample-specific factor matrices give a sample-specific operator.
\item \textbf{(P3) Each layer is a posterior inference step.} A PT forward pass is a finite sequence of MFVI iterations, each of which is a Bayesian coordinate-ascent update on the CRF. The latent state at any layer is interpretable as a (factorised) posterior over CRF labels, and the iteration schedule is itself a design choice.
\end{itemize}

These three primitives each opens a distinct programmability angle that is not naturally accessible to a Transformer treated as a neural black box, and each will, after we lift PT to ST-PT, give rise to its own Research Question.

\subsection{From PT to ST-PT: A Cornerstone for Time Series}

PT was originally designed for natural language processing, where sequences are one-dimensional and each token has intrinsic semantic content. Time series are different on two aspects: observations are multivariate, so a channel axis is essential as revealed in iTransformer~\cite{iTransformer}; a single time step carries almost no semantic meaning unless aggregated into patches as revealed in PatchTST~\cite{patchtst}. 

Therefore we extend PT to the \emph{Spatio-Temporal Probabilistic Transformer (ST-PT)}: a two-dimensional factor graph whose nodes sit at every (channel, time-patch) coordinate, whose ternary factors carry messages along both the temporal and the channel axes, and whose head variables compete over these two directions via a joint softmax. Defined over a 2D spatio-temporal factor graph, ST-PT preserves the expressive power of Transformer-style computation while inheriting all three programmable primitives P1--P3 from PT. The model is a shared framework---the \emph{cornerstone}---that we use for the following three research questions.

\subsection{Three Motivations, Three Research Questions}
\label{sec:rq}

Each of the three primitives in~\ref{3primitive} raises a concrete question about \emph{how to use} ST-PT in time series modelling, and each question is anchored in a different existing paradigm whose limitations the corresponding primitive is well-positioned to address. We label the three studies RQ1, RQ2, RQ3 and use these labels consistently throughout the report.

\begin{description}
\item[RQ1 --- Graph as the prior interface (built on P1).] Existing approaches inject domain priors into deep time-series models implicitly, through architectural choices: decomposition blocks~\cite{autoformer,fedformer}, sparse attention patterns~\cite{informer}, or task-specific spatio-temporal networks~\cite{dcrnn}. Each new type of prior demands a new architectural trick, with no shared interface~\cite{injectinductivebias,compare_prior}. Primitive P1 turns the factor graph itself into the prior interface, raising the question: \emph{can well-known symbolic time-series priors---e.g., periodicity, trend, lag---be injected into ST-PT through structural modifications of the factor graph (added nodes, modulated potentials, new pairwise edges) rather than through architectural hacks, and when does such an injection actually pay off?} We will conduct study on this question via controlled data-scarce and noise perturbed experiment setting. 

\item[RQ2 --- Factors as the condition interface (built on P2).] Existing conditional time series generators consume the external condition through \emph{feature-level} modulation of a fixed denoiser (adaptive LayerNorm, cross-attention, classifier-free guidance)~\cite{narasimhan2024timeweaver,gu2025verbalts,ge2025t2s,li2025bridge}: the network topology is unchanged, and only hidden activations are shifted or reweighted. Primitive P2 makes the factor matrices themselves the natural site at which a condition can act, raising the question: \emph{can the condition instead generate the factor matrices, so that each condition instantiates a different CRF---and hence a different message-passing operator---rather than merely modulating one?} We will instantiate this idea as the \textit{\textbf{P}robabilistic \textbf{T}ransformer \textbf{F}actor \textbf{G}enerating} (PT-FG) model and evaluate it on ConTSG-Bench~\cite{contsgbench2026} against other conditional generators.

\item[RQ3 --- MFVI as the latent-transition interface (built on P3).] Existing autoregressive forecasters, whether output-space~\cite{salinas2019deepvar,hochreiter1997lstm,lai2018lstnet} or latent-space~\cite{nextlat}, drive their step-to-step transition through a free-form learned module without explicit Bayesian semantics, and offer no built-in remedy for the exposure bias common to any AR rollout. Primitive P3 makes each ST-PT layer a Bayesian posterior update, raising the question: \emph{can MFVI on the extended ST-PT factor graph supply latent-space AR with a principled posterior update that fuses observation evidence, transition prior, cross-channel messages, etc? And can a CRF teacher (an encoder MFVI over the full history-plus-future sequence, available at training time) distill its latents into the AR student to counter exposure bias?} We evaluate the resulting model on classical long-horizon benchmarks against output-space AR baselines.
\end{description}

\subsection{Contributions and Positioning}

Our contribution is the exposition of ST-PT as a programmable framework and a coordinated empirical investigation of three uses of it under three distinct motivations. Concretely:
\begin{itemize}
\item We articulate ST-PT as a 2D spatio-temporal extension of PT and identify three programmable levers---graph topology and potentials, factor-matrix parameterization, and MFVI inference protocol---through which it becomes a programmable framework rather than a single model.
\item We implement and evaluate one study per Research Question: \textbf{RQ1} (graph-level priors for few-shot forecasting; synthetic and real data, with a noise-robustness analysis and an honest scoping result on real data); \textbf{RQ2} (condition-programmable factors PT-FG for conditional generation; 10 ConTSG benchmarks with a compositional-generalization ablation); and \textbf{RQ3} (MFVI-based latent-space AR forecasting; long-horizon benchmarks against AR baselines).
\end{itemize}

\section{Related Work}
\label{sec:related}
\paragraph{Probabilistic Transformer (PT)}
Our work is theoretically grounded in the Probabilistic Transformer (PT)~\cite{pt} in the context of natural language processing. PT establishes a fundamental mathematical equivalence between the Transformer's self-attention mechanism and Mean-Field Variational Inference (MFVI) in a Dependency Conditional Random Field (CRF). This insight reinterprets the Transformer not as a neural network, but as a structured probabilistic graphical model. While PT was originally designed to capture syntactic dependencies in text, we would like to extend its potential for time series modeling. We note that the name ``Probabilistic Transformer'' is also used by~\cite{pt4ts_nip} for a Transformer that outputs probabilistic (distributional) forecasts for time series; despite the shared terminology, that work is essentially unrelated to ours, since our foundation is the MFVI/CRF equivalence established by~\cite{pt}.

\paragraph{Multivariate Time Series Forecasting.}
Multivariate time series forecasting has evolved from Vector Autoregression and ARIMA~\cite{arima,stl} to a rich family of deep architectures~\cite{ts_dl}. Early Transformer variants focused on reducing the quadratic complexity of self-attention~\cite{logsparse,informer,pyraformer}, followed by decomposition-based approaches~\cite{autoformer,fedformer,etsformer,film}, patch-based tokenization~\cite{patchtst}, channel-as-token inversion~\cite{iTransformer}, and MLP-mixing-style models~\cite{tsmixer,timemixer++}. Despite their sophistication, these models treat the network as a fixed black box and do not provide an explicit interface for injecting structured prior knowledge. The empirical dominance of the linear baseline DLinear~\cite{dlinear} and the theoretical critique in~\cite{transformerfail} suggest that raw expressive power is not a substitute for an appropriate inductive bias.

\paragraph{Time Series Forecasting under Data Scarcity.}
Under data scarcity, the dominant strategies compensate with \emph{external data}. Meta-learning adapts MAML-style objectives to temporal tasks~\cite{maml,fewshot}, and time-series foundation models such as TimeGPT~\cite{timegpt}, large pretrained LM-repurposing approaches~\cite{timellm,onefitsall}, and in-context fine-tuning~\cite{finetuning} leverage massive source corpora. These paradigms assume statistical overlap between source and target; under pronounced domain shift their benefits diminish. A complementary and less---explored direction is to compensate with \emph{structural human knowledge} instead of external data---precisely the setting in which the prior-injection capability of ST-PT (RQ1) is most relevant~\cite{fewshot_biomedical}.

\paragraph{Injecting Prior Knowledge into Time Series Models.}
Prior-knowledge injection has followed several disconnected threads. Physics-informed neural networks enforce governing equations as soft loss constraints~\cite{physicsprior}. Domain-specific models fuse physical mechanisms with statistical backbones, as in heat demand prediction~\cite{heatprediction}. Graph-based models inject spatial topology as a hard inductive bias, exemplified by DCRNN for traffic forecasting~\cite{dcrnn}. These approaches are effective in their niches but share no theoretical interface~\cite{injectinductivebias}. Crucially,~\cite{compare_prior} observed that na\"ive hard-coding often \emph{underperforms} learned representations inside Transformers, motivating a principled integration mechanism of the kind ST-PT offers.

\paragraph{Conditional Time Series Generation.}
Conditional time series generation (ConTSG) has progressed from class-conditioned GAN/VAE baselines such as TTS-CGAN~\cite{li2022ttscgan} and TimeVQVAE~\cite{lee2023timevqvae} to attribute-conditioned diffusion and transformer denoisers including TimeWeaver~\cite{narasimhan2024timeweaver}, WaveStitch~\cite{shankar2025wavestitch}, and TEdit~\cite{jing2024tedit}, and further to text-conditioned generation including VerbalTS~\cite{gu2025verbalts}, T2S~\cite{ge2025t2s}, BRIDGE~\cite{li2025bridge}, and DiffuSETS~\cite{lai2025diffusets}. Almost all these works inject the condition through \emph{feature-level modulation} of a fixed denoiser (adaptive LayerNorm, cross-attention, classifier-free guidance). Our RQ2 study departs from this line by using the condition to instantiate the denoiser's factor matrices, so that different conditions yield different CRFs. A recent benchmark ConTSG-Bench~\cite{contsgbench2026} provides standardized head-to-head evaluation across these methods.

\paragraph{Autoregressive Forecasting and Latent-Space AR.}
Autoregressive forecasters split into two camps. \emph{Output-space AR}---probabilistic models such as DeepVAR~\cite{salinas2019deepvar} and recurrent alternatives such as LSTM~\cite{hochreiter1997lstm} and LSTNet~\cite{lai2018lstnet}---regresses or samples a concrete next-step value or patch at every step. Thus, the per-step latent is effectively collapsed into the output space and inter-step information flow is mediated solely by whatever the recurrent state retains; resolution is bounded by the patch length and the cross-step bandwidth is bounded by the recurrent hidden size. Models inspired by NextLat~\cite{nextlat}, a \emph{Latent-space AR} paradigm, attempt to predict in a learned latent space to mitigate these resolution and bandwidth bottlenecks. However, their latent transitions are typically parameterized as opaque, free-form neural modules rather than structurally explicit Bayesian inference steps. While these latents may theoretically approximate belief states, the lack of an explicit inference mechanism leaves no principled computational site at which to organically fuse cross-channel evidence or external priors during rollout. Our RQ3 study replaces the latent transition with posterior inference on the ST-PT factor graph and uses a full-sequence CRF teacher to distill oracle latents against exposure bias.

\section{Preliminaries: PT and the ST-PT Cornerstone}
\label{sec:method}

\subsection{Problem Formulation}

Given a multivariate time series $\mathbf{X}=\{\mathbf{x}_1,\dots,\mathbf{x}_T\}\in\mathbb{R}^{T\times N}$ with $T$ time steps and $N$ channels, the forecasting task predicts the next $S$ steps $\mathbf{Y}=\{\mathbf{x}_{T+1},\dots,\mathbf{x}_{T+S}\}\in\mathbb{R}^{S\times N}$. The conditional generation task further assumes a condition vector $\mathbf{c}$ and asks for generating samples whose distribution is consistent with the data and with $\mathbf{c}$. The autoregressive forecasting task specializes the forecasting problem by producing $\mathbf{Y}$ one patch at a time and \emph{recursively}: at step $k$ the model feeds its own previously emitted predictions $\hat{\mathbf{x}}_{T+1},\dots,\hat{\mathbf{x}}_{T+k-1}$ (or, in latent-space variants, their latent encodings) back as input to infer the next patch $\hat{\mathbf{x}}_{T+k}$.

\subsection{Probabilistic Transformer, Revisited}

PT~\cite{pt} describes computation as MFVI on a CRF with three groups of latent variables: token-level discrete labels $\{Z_i\}$, per-head dependency variables $\{H_i^{(c)}\}$, and per-token global variables $\{G_i\}$. Three families of potentials define the CRF: unary potentials $\phi_u(Z_i)$ that couple $Z_i$ to local evidence; ternary potentials $\phi_t^{(c)}(H_i^{(c)},Z_i,Z_j)$ that couple $H_i^{(c)}$ to a candidate target $Z_j$; and binary potentials $\phi_b(Z_i,G_i)$ that couple $Z_i$ to a global topic $G_i$. Under MFVI, the $Z,H,G$ marginals are approximated by factorized distributions $q(Z),q(H),q(G)$, and iterative coordinate-ascent updates take the following form (writing $\psi_\star\!=\!\log\phi_\star$ for the log-potentials):
\begin{align}
q(H_i^{(c)}\!=\!j) &\;\propto\; \exp\!\big(F_i^{(c)}(j)\big),
   \label{eq:H}\\[-1pt]
\text{with}\quad F_i^{(c)}(j) &\;=\; \sum_{a,b} q(Z_i\!=\!a)\,q(Z_j\!=\!b)\,\psi_t^{(c)}(j,a,b),
   \label{eq:F}\\[2pt]
q(Z_i\!=\!a) &\;\propto\; \phi_u(Z_i\!=\!a)\,\exp\!\big(G_i(a)\big)\prod_g\phi_b(a,g)^{q(G_i=g)},
   \label{eq:Z}\\[-1pt]
\text{with}\quad G_i(a) &\;=\; \sum_c\sum_j q(H_i^{(c)}\!=\!j)\sum_b q(Z_j\!=\!b)\,\psi_t^{(c)}(j,a,b).
   \label{eq:G}
\end{align}
Equations~\eqref{eq:F} and~\eqref{eq:G} are the two messages that cross each ternary factor in one MFVI iteration: the \emph{forward} message $F$ (from $Z$ to $H$) scores every candidate parent $j$ by aggregating pairwise compatibility under the current $Z$ beliefs, and softmax normalization of $F$ over $j$ produces $q(H)$; the \emph{backward} message $G$ (from the updated $H$ back to $Z$) collects the weighted ternary-factor evidence that each position should absorb into its $Z$-belief. The unary potential $\phi_u$ supplies local evidence, and the binary potential $\phi_b$ contributes a $q(G_i)$-weighted coupling to the global topic variable. \cite{pt} shows that with a low-rank tensor decomposition of $\psi_t^{(c)}$, equation~\eqref{eq:F} takes the form of a scaled dot-product attention score, softmax over $j$ in equation~\eqref{eq:H} reproduces the attention weights, and the $Z$ update in equation~\eqref{eq:Z}---together with $G$ from equation~\eqref{eq:G}---reproduces the feed-forward block of a Transformer. Damping the update of $q(Z)$ simulates a residual connection. The forward pass of a Transformer is therefore a fully specified message-passing procedure on a CRF whose factor matrices are learned jointly with the task objective.

\subsection{Why PT Does Not Directly Fit Multivariate Time Series}

Two gaps prevent a na\"ive application of PT to multivariate time series:
\begin{enumerate}
\item \emph{No channel axis.} PT was designed for one-dimensional token sequences. Multivariate series have $N$ channels whose inter-dependencies are themselves a central modelling target~\cite{iTransformer}.
\item \emph{Weak per-token semantics.} A single time step is not analogous to a word---it carries almost no semantic content in isolation~\cite{patchtst}.
\end{enumerate}
\begin{figure}[t]
    \centering
    \includegraphics[width=\textwidth]{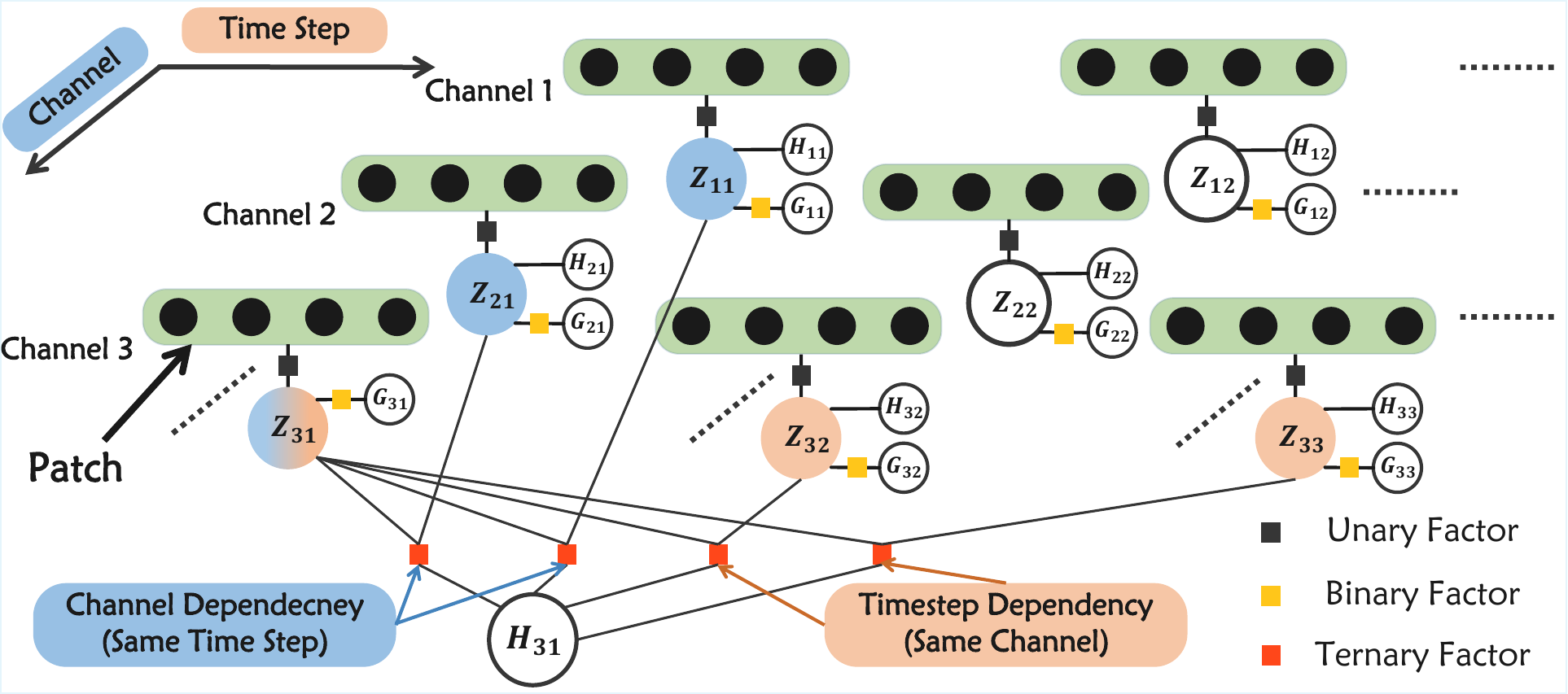}
    \caption{Structure of ST-PT: a 2D factor graph on (channel $i$, time-patch
  $t$) coordinates with $Z/H/G$ latent variables, unary/ternary/binary
  factors, and a joint softmax over temporal and cross-channel dependencies.}
    \label{fig:st-pt}
\end{figure}
\subsection{ST-PT Cornerstone Model}

ST-PT addresses all these gaps by lifting PT's factor graph to two dimensions. The structure of ST-PT is presented in Figure \ref{fig:st-pt}. The model consists of the following components and designs:

\paragraph{Unary Factor and $Z$ variables}
The input raw data will be Z-normalized and then sliced into small patches. For one patch consisting of $ps$ observed data point, there will be a unary factor expressed by a MLP. The Z node connecting the patch can be therefore somehow regarded as an embedding. After iterations of information update, Z variables will be feed into a simple MLP to predict the future time series sequence.

\paragraph{Binary Factor and $G$ variables}
In original Probability Transformer paper, the variant of vanilla PT adds $G$ variables and binary factors to simulate FFN and is adopted here. The binary factor here is a matrix and will be used in the matrix multiplication in the information conveyance formula. Note that one $Z$ variable's $H$ only connects to those $Z$ variables who either share the same channel or timestep, and doesn't connect to itself. 

\paragraph{Ternary Fector and $H$ variables}
$H$ variables models the dependency relationship, and ternary factor is used to quantify and evaluate the dependency. In our implementation, we have two ternary factor matrices, one for the same-channel patches dependency and one for the same-timestep patches dependency. 

Let $p$ be the patch length and $P=T/p$ the number of patches along the time axis. The input is reshaped into $\mathbf{X}=\{\mathbf{x}_{i,t}\}\in\mathbb{R}^{N\times P\times p}$ with $\mathbf{x}_{i,t}\in\mathbb{R}^{p}$. A learnable unary map $\phi_u$ embeds each patch into a $d$-dimensional hidden state. Patches restore local semantic content and reduce complexity, following PatchTST~\cite{patchtst}.

ST-PT places one $Z$-node, one $H$-node, and one $G$-node at every coordinate $(i,t)$ where $i\in[N]$ is a channel and $t\in[P]$ is a patch index. Each $H_{i,t}^{(c)}$ may point either to a same-channel patch (temporal dependency) or to a same-timestep patch in a different channel (cross-channel dependency). Two disjoint sets of ternary potentials are instantiated, $(U^{\text{time}},V^{\text{time}})$ and $(U^{\text{chan}},V^{\text{chan}})$, yielding respective message-$F$ scores $F^{\text{time}}_{i,t,c}(s)$ and $F^{\text{chan}}_{i,t,c}(j)$. Distinct from designs that alternate between time and channel mixing, like TSMixer~\cite{tsmixer} and MLPMixer~\cite{mlpmixer}, ST-PT normalizes temporal and channel candidates \emph{jointly}:
\begin{equation}
q\!\big(H_{i,t}^{(c)}\big) \;=\; \mathrm{softmax}\!\Big(\tfrac{1}{\lambda_H}\big[F^{\text{time}}_{i,t,c}(\cdot),\,F^{\text{chan}}_{i,t,c}(\cdot)\big]\Big),
\end{equation}
so that the two kinds of dependency compete on equal footing within a single message. The resulting messages $G^{\text{time}}$ and $G^{\text{chan}}$ are then fed into the $Z$ update together with the global topic message $G^{\text{binary}}$ from $\phi_b$ and the unary $\phi_u(\mathbf{x}_{i,t})$.

Rotary position embeddings (RoPE) are applied along both axes, with separate parameter sets for the time and channel directions. This distinguishes a temporal shift from a channel shift without coupling them into a single one-dimensional position. The encoder iterates $K$ rounds of the MFVI-style update with damping. The final $Z$ beliefs are projected through an MLP head to produce the forecasting (or denoising) output. This cornerstone model is the backbone used in all three RQs below.

\section{Three Research Questions: ST-PT as a Programmable Framework}

\begin{figure}[t]
    \centering
    \includegraphics[width=\textwidth]{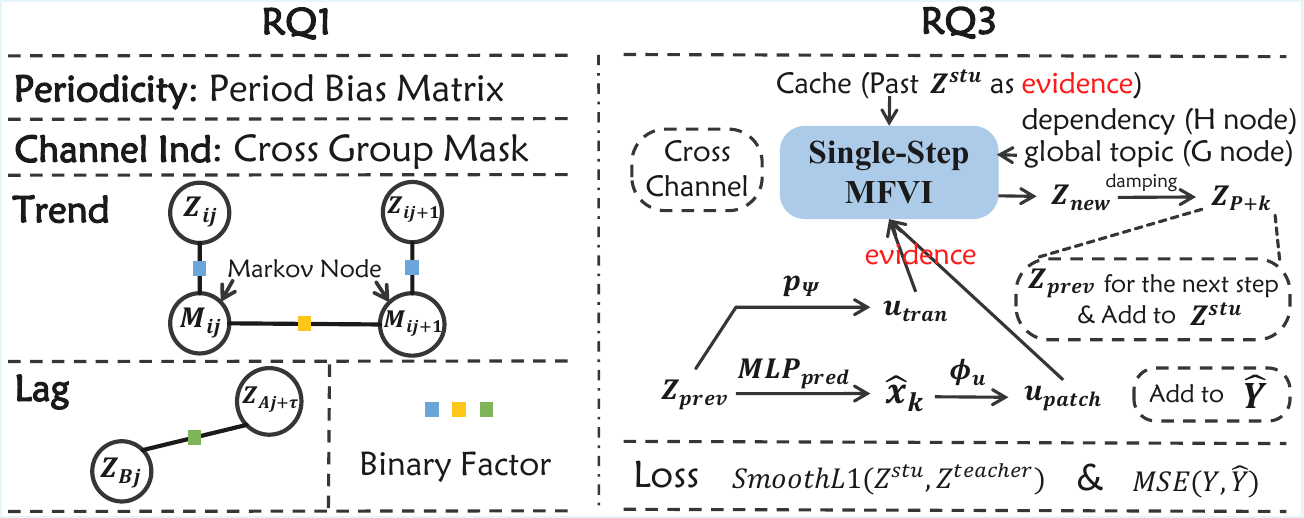}
\caption{(a) Three graph-level mechanisms for injecting symbolic priors
  into ST-PT's factor graph (RQ1, Section~\ref{sec:rq1}): \emph{periodicity}
  multiplies the temporal message-$F$ logits by a per-channel cosine
  similarity matrix (potential modulation); \emph{trend} augments the graph
  with a per-channel latent HMM chain coupled to $Z$ (node augmentation);
  \emph{lag} adds new bilinear pairwise edges between causally linked
  channels at a known temporal offset (factor engineering). (b) Computation
  flow of the RQ3 latent-space AR model (Section~\ref{sec:rq3},
  Appendix~\ref{app:latent-ar-pseudo}): a causal encoder MFVI turns the
  history patches into a latent cache, and each AR step infers the
  next-slice latent by a single-step MFVI update that fuses a dual-pathway
  unary (re-encoded prediction + transition prior), temporal messages from
  the cache, cross-channel messages among the simultaneous new latents, and
  a global topic.}
    \label{fig:rq13}
\end{figure}

The three RQs introduced in Section~\ref{sec:rq} share a single technical theme: ST-PT's three distinct levers---\emph{graph}, \emph{parameters}, and \emph{inference protocol}---each unlock a modelling regime that opaque Transformers do not cleanly support. Each RQ is matched with one study experiment in Section~\ref{sec:experiments}.

\subsection{RQ1: Graph-level Prior Injection}
\label{sec:rq1}

\paragraph{Claim.} The factor graph of ST-PT is itself the interface for injecting symbolic priors. Domain knowledge can be expressed by structural modifications to the graph, yielding energy-landscape shaping that is principled rather than ad-hoc. Here we take four forms of prior as study subject.

\paragraph{Four priors as study subjects.} We select four canonical time-series priors---\emph{periodicity}, \emph{trend}, \emph{lag}, and \emph{channel-independence}---as the study subjects of RQ1, and realise each one as a concrete graph-level modification of the ST-PT factor graph, as illustrated in Figure~\ref{fig:rq13}. Each prior has its own modelling mechanism, acting on a different site of the graph. The detailed implementation of these priors can be referred to Appendix~\ref{app:prior}.

\begin{itemize}
\item \emph{Potential modulation (periodicity prior).} For each channel $i$ we construct a per-channel period bias matrix $\mathbf{P}_i\in\mathbb{R}^{P\times P}$ whose entries encode cosine similarity between patches under the known period(s) of that channel. The temporal message $F^{\text{time}}$ is multiplied element-wise by $\mathbf{P}_i$, biasing the message-passing to respect the known periodicities. No new nodes or factors are added.
\item \emph{Node augmentation (trend prior).} We introduce a second layer of latent nodes $M_{i,t}$, one per $(i,t)$, with a Markovian pairwise potential and an observation potential that ties $M_{i,t}$ to $Z_{i,t}$. The resulting extended graph behaves like an explicit hidden Markov model coupled to the original ST-PT, and injects a smooth-trend inductive bias.
\item \emph{Factor engineering (lag prior).} For each pair of causally-related channels $(A,B)$ with known lag $\tau$ we add a new pairwise potential $\phi_{AB}(Z_{A,t},Z_{B,t+\tau})$, implemented as a learnable bilinear form. This adds a new edge class to the factor graph, carrying messages across channels with a temporal offset, without disturbing the existing time and channel ternary factors.
\item \emph{Edge masking (channel-independence prior).} Given a declared channel-group partition $\mathcal{G}=\{G_1,\dots,G_K\}$, we mask the cross-channel message $F^{\text{chan}}$ so that only intra-group messages survive, i.e.\ the cross-group entries are set to $-\infty$ before the joint softmax. In graph terms, this amounts simply to deleting a subset of cross-channel edges emanating from the $H$-nodes, namely those that would otherwise connect $H$-nodes belonging to different groups. The modification acts on the support of an existing message rather than introducing new structure: admissible cross-channel edges are kept untouched while cross-group edges are forbidden outright, encoding a hard independence assumption between the declared channel groups.

\end{itemize}

\paragraph{Expected regime.} Under Bayesian intuition, each prior's pay-off depends on whether its structural assumption carries information the data cannot already supply on its own: a prior helps most when training samples are scarce, when noise dominates, or when its assumption matches the dominant signal, and its marginal contribution is diluted when the data is abundant and already exhibits multiple overlapping patterns. This expectation applies directly to the periodicity, trend, and lag priors, each of which inscribes a specific structural assumption into the factor graph. The channel-independence prior sits in a subtly different regime: because its modification only \emph{forbids} cross-group $H$-edges that the unrestricted ternary attention is otherwise free to pass, its benefit exists only when the data-driven attention would in fact allocate non-trivial mass to those now-forbidden edges; if the data itself already down-weights cross-group links, the mask is redundant. We report this case (Section~\ref{sec:expA}) to delineate what kind of prior ST-PT's graph interface does and does not benefit from.

\subsection{RQ2: Condition-Programmable Factors}
\label{sec:rq2}

\paragraph{Claim.} ST-PT's factor matrices are not auxiliary hyperparameters; they \emph{are} the CRF's potentials. By letting an external condition generate these matrices on a per-sample basis, we obtain a structural form of conditioning whose effect is to instantiate a different CRF---and hence a different message-passing operator---for each condition. This modeling philosophy contrasts with the feature-level modulation (adaptive LayerNorm, cross-attention, classifier-free guidance) used by existing conditional diffusion time-series generators~\cite{gu2025verbalts,ge2025t2s,li2025bridge,shankar2025wavestitch} and can be similar with Hypernet~\cite{hypernet}.

\paragraph{Factor generator design (PT-FG).} The condition is first encoded into a hidden control vector $\mathbf{h}$ via
\begin{equation}
\mathbf{h} \;=\; \mathrm{MLP}\!\big(\mathbf{W}_c\mathbf{c} + \mathrm{TimeEmb}(t)\big),
\end{equation}
where $t$ is the diffusion timestep. $\mathbf{h}$ then drives three generators:
\begin{description}
\item[Unary conditioner.] Produces patch-wise scale, shift, and segment gate vectors $(\mathbf{s},\mathbf{b},\mathbf{g}_{\mathrm{seg}})$ from $\mathbf{h}$, yielding an adaptive LayerNorm-style modulation of the initial $Z$ states.
\item[Ternary basis generator.] For each of $\{U^{\text{time}},V^{\text{time}},U^{\text{chan}},V^{\text{chan}}\}$ we maintain a shared base matrix $\mathbf{U}_0$ plus $K$ learnable basis matrices $\{\mathbf{B}_k\}$. The condition predicts coefficients $\alpha_k(\mathbf{h})$ and row/column scales, yielding $\mathbf{U}(\mathbf{h})=(\mathbf{U}_0+\sum_k\alpha_k(\mathbf{h})\mathbf{B}_k)\odot\mathbf{r}(\mathbf{h})\odot\mathbf{c}(\mathbf{h})^\top$. Basis mixing is far more parameter-efficient than a full hypernetwork.
\item[Binary topic generator.] A similar basis-decomposition generator produces the global topic factor $\mathbf{W}(\mathbf{h})$.
\end{description}

\paragraph{Diffusion training.} The condition-generated factors define a sample-specific ST-PT denoiser $\epsilon_\theta(\mathbf{x}_t,t,\mathbf{c})$. Training minimizes the standard noise-prediction loss; classifier-free guidance is supported through null-condition dropout. Crucially, the condition programs \emph{which} CRF is doing the denoising, not merely \emph{where} the denoiser attends.

\paragraph{Why this matters for compositional generalization.} Consider controllable generation over multiple structured attributes. In feature-modulation approaches, the interaction among attributes is mediated solely by hidden activations and can easily memorize training co-occurrences. Under the CRF view, the factors act as the operator of an inference procedure; if each attribute independently contributes basis coefficients and scales, a novel combination of attributes naturally synthesize a novel operator. The empirical study in Section~\ref{sec:expB} confirms that enforcing independent attribute encoding (as opposed to multi-head cross-talk among attribute embeddings) improves compositional generalization but still underperforms text condition encoding which is consistent with the observation in ConTSG-Bench~\cite{contsgbench2026}.

\subsection{RQ3: MFVI-based Latent-Space Autoregressive Forecasting}
\label{sec:rq3}

\paragraph{Claim.} Output-space AR forecasters (e.g.,\ DeepVAR~\cite{salinas2019deepvar}, LSTM-AR~\cite{hochreiter1997lstm}, LSTNet~\cite{lai2018lstnet}) commit to a concrete next-step value or patch at every step, so cross-step information flow is mediated entirely by whatever the recurrent state retains. Latent-space AR~\cite{nextlat} decouples rollout from per-step output commitment, but its latent transition is a free-form learned module without explicit Bayesian semantics. ST-PT instead makes the latent update an explicit \emph{posterior inference step} on an extended factor graph, fusing observation evidence, transition prior, temporal cache, cross-channel messages and a global topic through principled message passing, and uses a full-sequence CRF teacher to distil oracle latents into the AR student.

\paragraph{Design intuition.}
Three modelling choices do the work, each pinned to a specific expected effect. \emph{(i) A posterior-inference latent transition.} Replacing a free-form transition module with one MFVI iteration forces the rolled-out latent to live in the same belief space as the encoder's latents, so the long-range temporal, cross-channel, and topic structure already captured by the encoder is reused directly at rollout time instead of being re-learned by a bespoke transition module. \emph{(ii) A dual-pathway unary.} Splitting the new slice's unary into a transition-prior term and an observation-likelihood term supplies the ingredient pair that any posterior update requires; each term alone has a specific failure mode (discussed below) that the other term corrects. \emph{(iii) A CRF teacher built from the same operator.} A full-sequence encoder MFVI is available at training time and lives in the student's own belief space, so it provides a same-operator oracle that can directly supervise the student's \emph{rolled-out latents}, not only its outputs---without a second model or a learned alignment. 

\paragraph{Dual-pathway evidence.} At each AR step $k$ we infer a new latent $z_{P+k+1}$ by running a single-step MFVI iteration. The ``unary'' for the new slice is the sum of two pathways:
\begin{itemize}
\item \emph{(i) Patch unary.} The predicted patch $\hat{\mathbf{x}}_{P+k}=\mathrm{MLP}_{\text{pred}}(z_{P+k})$ is re-encoded through the unary map $\phi_u$. This grounds the AR step in the observation space and forces commitment.
\item \emph{(ii) Transition prior.} A learned MLP $p_\psi(z_{P+k})$ generates a latent-space shortcut. Unlike patch unary, it bypasses the patch-length bottleneck and preserves high-dimensional information.
\end{itemize}
Both evidence sources enter the MFVI update together with: (iii) temporal messages from cached past latents (the $U^{\text{time}}$ factors act on a causal cache that includes all $P$ observed patches and all previously inferred future patches); (iv) cross-channel messages from the simultaneous latents at other channels; and (v) the global topic message.

\emph{Why two pathways.} The patch unary passes through a $d\!\to\!p\!\to\!d$ projection ($d\!=\!256$, $p\!=\!8$), so anything absent from the $p$-dimensional patch is dropped at every step; using only this pathway would shrink the effective cross-step information bandwidth to $p$ floats and cause fine-grained latent context to be forgotten after one step. The transition prior avoids this bottleneck but is entirely internal, with no per-step tie to a realisable observation, so using only this pathway leaves nothing that prevents the rollout from drifting into a self-consistent but data-inconsistent region of latent space. Combining them realises a standard \emph{prior + likelihood} pair: the transition prior is the dynamic prior on the next latent, the patch unary is the observation-level likelihood, and MFVI fuses them into a posterior. The other three message sources (cache, cross-channel, topic) are context for this update---the same role they play inside the encoder.

\paragraph{CRF teacher distillation.}
During training the full history-plus-ground-truth-future sequence is available; we run a parallel MFVI on the \emph{full} sequence to obtain oracle posteriors $z^{\mathrm{teacher}}_{P+1:P+S_f}$. Crucially, the RQ3 encoder MFVI is \emph{causal} along the time axis (position $t$ attends only to $s\!\le\!t$, implemented as an additive $-\infty$ mask on the upper triangle of the time-axis message-$F$ logits); since both teacher and student use the same causal operator, their history-position latents ($t\!<\!P$) are \emph{identical}, and the two paths can differ only at the $P_f$ future positions---where the teacher has ground-truth patch-unary evidence while the student has only its self-rolled predictions. The distillation loss therefore acts only on the future positions, pulling the student's self-rolled posterior toward the teacher's oracle-informed one; in practice we use either a KL-divergence-like term under the SquaredSoftmax normalization or a SmoothL1/cosine combination. Because the teacher and the student share the same MFVI operator and the same factor matrices, adn the distillation amounts to telling the student what its own posterior would have been if it could see the future, which can be regarded as a principled antidote to cumulative error without introducing a qualitatively different network at training time.

\emph{Why this is a natural training signal.} Because the teacher and the student are produced by the \emph{same} MFVI operator---differing only in that the teacher's input includes the future---their latents live in the same belief space, so a direct element-wise match between them is already commensurable and needs no alignment layer. Operationally this places the supervision on the student's \emph{beliefs} rather than on its inputs or outputs: the student's inputs remain self-rolled at training time (unlike teacher forcing, which aligns inputs but leaves a train/test input-distribution mismatch), while each rolled-out latent is pulled toward what the student's own operator would produce with future visibility---attacking exposure bias from the posterior side rather than the input side.

\paragraph{Model and computation flow.}
Concretely, the RQ3 model (full pseudocode in Appendix~\ref{app:latent-ar-pseudo}) reuses the ST-PT ternary and binary factors of Section~\ref{sec:method} and deploys them in three coordinated phases. \emph{Phase 1 --- encoder MFVI on history.} The history patches are embedded by the shared unary $\phi_u$ and passed through $K_{\text{enc}}$ rounds of a \emph{causal} MFVI (a $-\infty$ mask on the upper triangle of the time-axis message-$F$ logits), yielding a latent cache $\mathbf{Z}^{\text{stu}}\!\in\!\mathbb{R}^{N\times P\times d}$ that summarises every history position under the same 2D factor graph as the cornerstone encoder. \emph{Phase 2 --- AR rollout.} For $k\!=\!1,\dots,P_f$ the model (a) reads the most recent cached latent $z_{P+k-1}$ and emits the next patch $\hat{\mathbf{x}}_{k}\!=\!\mathrm{MLP}_{\text{pred}}(z_{P+k-1})$; (b) forms the new slice's unary as the dual-pathway sum $u_{\text{patch}}\!+\!u_{\text{trans}}$ with $u_{\text{patch}}\!=\!\phi_u(\hat{\mathbf{x}}_{k})$ and $u_{\text{trans}}\!=\!p_\psi(z_{P+k-1})$; (c) runs a $\mathrm{SingleStepMFVI}$---one round of the same causal update restricted to the single new time slice---which fuses this unary with temporal messages from the cache, cross-channel messages from the simultaneous new latents at other channels, and the global topic message, producing $z_{\text{new}}$; (d) damps $z_{P+k}\!=\!\alpha\,z_{P+k-1}+(1-\alpha)\,z_{\text{new}}$ and appends it to the cache, so that the next step reads it as context (closing the AR loop on the latent side). \emph{Phase 3 (training only) --- CRF teacher.} On the concatenated sequence $[\mathbf{X}_{\text{hist}}\,\|\,\mathbf{Y}]$, a parallel encoder MFVI with the same causal operator and the same factor matrices produces teacher latents $\mathbf{Z}^{\text{teacher}}$; because the operator is causal, the teacher's and student's history-position latents coincide exactly, so the distillation loss acts only on the $P_f$ future positions. The total training objective is $\mathrm{MSE}(\hat{\mathbf{Y}},\mathbf{Y})+\lambda_{\text{latent}}\,\mathcal{L}_{\text{distill}}$. At inference only Phases~1--2 run; Figure~\ref{fig:rq13}(b) illustrates the overall computation flow.

\paragraph{What is new compared to latent AR in general.} Three structural differences are worth highlighting:
\begin{enumerate}
\item The latent transition is not a stand-alone MLP but a fully specified Bayesian update on a factor graph with rich temporal and cross-channel structure.
\item Training and inference share the MFVI operator; the teacher latent is reached by the same dynamics as the student, merely with different evidence.
\item The approach carries CRF semantics into the latent space: the latent is not an arbitrary continuous vector but a (softened) probability distribution over CRF labels, which gives the distillation loss a clear meaning.
\end{enumerate}

\section{Experiments}
\label{sec:experiments}

\subsection{Cornerstone Validation: ST-PT for Long-Term Forecasting}
\label{sec:exp-cornerstone}

Before activating any RQ-specific lever, we validate that the bare ST-PT cornerstone is itself a competitive long-term forecaster, so that the RQ1--RQ3 gains and losses reported below are differential effects on top of a reasonable backbone rather than artefacts of an under-trained one. We evaluate ST-PT on the standard 7-dataset long-term forecasting suite (Weather, Electricity, Traffic, ETTh1, ETTh2, ETTm1, ETTm2) at horizons $\{96,192,336,720\}$, against strong 2023-era baselines: TSMixer~\cite{tsmixer}, TimeMixer~\cite{timemixer++}, PatchTST~\cite{patchtst}, TimesNet~\cite{timesnet}, and Crossformer~\cite{crossformer}. Table~\ref{tab:cornerstone-avg} reports Test MSE\,/\,MAE averaged over the four horizons; the full per-horizon table and experimental setting are in Appendix~\ref{app:cornerstone-full}.

\begin{table}[h]
\centering
\small
\setlength{\tabcolsep}{4pt}
\caption{Cornerstone validation: long-term forecasting on seven benchmarks, Test MSE / MAE averaged over four horizons $\{96,192,336,720\}$ (lower is better). The bare ST-PT cornerstone is broadly comparable to strong 2023 baselines on every benchmark, without any RQ-specific lever activated. Per-horizon raw numbers and the experimental setting are in Appendix~\ref{app:cornerstone-full}.}
\label{tab:cornerstone-avg}
\begin{tabular}{lcc cc cc cc cc cc}
\toprule
\multirow{2}{*}{Dataset} & \multicolumn{2}{c}{ST-PT (Ours)} & \multicolumn{2}{c}{TSMixer} & \multicolumn{2}{c}{TimeMixer} & \multicolumn{2}{c}{PatchTST} & \multicolumn{2}{c}{TimesNet} & \multicolumn{2}{c}{Crossformer} \\
\cmidrule(lr){2-3}\cmidrule(lr){4-5}\cmidrule(lr){6-7}\cmidrule(lr){8-9}\cmidrule(lr){10-11}\cmidrule(lr){12-13}
& MSE & MAE & MSE & MAE & MSE & MAE & MSE & MAE & MSE & MAE & MSE & MAE \\
\midrule
Weather     & 0.245 & 0.276 & 0.225 & 0.264 & 0.222 & 0.262 & 0.241 & 0.264 & 0.251 & 0.294 & 0.380 & 0.442 \\
Electricity & 0.177 & 0.273 & 0.160 & 0.257 & 0.156 & 0.247 & 0.159 & 0.253 & 0.193 & 0.304 & 0.186 & 0.283 \\
Traffic     & 0.446 & 0.289 & 0.408 & 0.284 & 0.388 & 0.263 & 0.391 & 0.264 & 0.620 & 0.336 & 0.542 & 0.283 \\
ETTh1       & 0.437 & 0.434 & 0.412 & 0.428 & 0.411 & 0.423 & 0.413 & 0.434 & 0.495 & 0.450 & 0.600 & 0.557 \\
ETTh2       & 0.368 & 0.399 & 0.355 & 0.401 & 0.316 & 0.384 & 0.324 & 0.381 & 0.371 & 0.404 & 0.564 & 0.548 \\
ETTm1       & 0.393 & 0.403 & 0.347 & 0.375 & 0.348 & 0.376 & 0.353 & 0.382 & 0.400 & 0.406 & 0.514 & 0.511 \\
ETTm2       & 0.283 & 0.332 & 0.267 & 0.322 & 0.256 & 0.316 & 0.256 & 0.317 & 0.291 & 0.333 & 0.621 & 0.510 \\
\bottomrule
\end{tabular}
\end{table}

On all seven benchmarks the bare ST-PT cornerstone tracks the 2023 strong baselines within a few percent on both MSE and MAE, never lagging any single dataset by a large margin. We therefore treat the cornerstone as a reasonable starting backbone for the three RQ-specific studies that follow.

\subsection{RQ1: Priors for Few-Shot and Noise Regimes}
\label{sec:expA}

\paragraph{Setup.}
We construct three synthetic datasets (Lag, Periodicity, Trend) each instantiated at three sample sizes $N\in\{150,1500,15000\}$. Lag data has $\mathrm{enc\_in}=6$ with paired channels $(0,1),(2,3),(4,5)$ and known lag $\tau=8$; Periodicity and Trend data have $\mathrm{enc\_in}=10$ and $\mathrm{seq\_len}=\mathrm{pred\_len}=96$ as in standard long-term forecasting benchmarks. Baselines are DLinear~\cite{dlinear}, iTransformer~\cite{iTransformer}, PatchTST~\cite{patchtst}, and the vanilla ST-PT cornerstone. The three additive prior variants (ST-PT+Lag, ST-PT+Period, ST-PT+Trend) follow the graph modifications described in Section~\ref{sec:rq1}; the restrictive channel-independence probe (ST-PT+Indep) is evaluated on the Lag task only, where the ground-truth channel partition is known. Full construction details for the three synthetic datasets are given in Appendix~\ref{app:syn-data} and more details on the settings of following experiments can be referred to Appendix~\ref{app:impl}.

\paragraph{Synthetic few-shot results (Tables~\ref{tab:synth-fewshot} and~\ref{tab:synth-indep}).}
On the three synthetic tasks, all three \emph{additive} prior-injected ST-PT variants beat the vanilla ST-PT cornerstone in the data-scarce $N\!=\!150$ regime, and the advantage gradually shrinks as $N$ grows---exactly what Bayesian intuition predicts once data compensates for the missing prior. Classical Transformer baselines (iTransformer) fail catastrophically at $N\!=\!150$; DLinear is competitive but is dominated by ST-PT on every task. The fourth, \emph{restrictive} channel-independence probe, however, produces essentially no change over the vanilla cornerstone on Lag data at any sample size (Table~\ref{tab:synth-indep}): even though the masked partition exactly matches the ground-truth channel pairs, enforcing the mask neither helps nor hurts, indicating that the unrestricted ternary attention already concentrates cross-channel messages inside the causal pairs on its own and the mask is simply redundant. The synthetic side of RQ1 therefore gives a clean three-success-plus-one-failure picture: priors that \emph{add} matched structure help when data is scarce; a prior that only \emph{removes} freedom does not help when the data-driven model already imposes the same restriction.

\begin{table}[t]
\centering
\small
\caption{Few-shot synthetic forecasting (Test MSE, lower is better). Prior-injected ST-PT variants are strongest at $N=150$, with the advantage shrinking as $N$ grows.}
\label{tab:synth-fewshot}
\begin{tabular}{llccccc}
\toprule
Task & $N$ & ST-PT & ST-PT+Prior & iTransformer & PatchTST & DLinear \\
\midrule
\multirow{3}{*}{Lag}
& 150    & 0.261 & \textbf{0.219} & 0.654 & 0.295 & 0.465 \\
& 1500   & 0.021 & \textbf{0.019} & 0.235 & 0.049 & 0.167 \\
& 15000  & 0.0026& \textbf{0.0023}& 0.235 & 0.009 & 0.027 \\
\midrule
\multirow{3}{*}{Periodicity}
& 150    & 0.230 & \textbf{0.222} & 0.580 & 0.250 & 0.455 \\
& 1500   & 0.216 & \textbf{0.213} & 0.221 & 0.238 & 0.209 \\
& 15000  & \textbf{0.194} & 0.195 & 0.193 & 0.201 & 0.201 \\
\midrule
\multirow{3}{*}{Trend}
& 150    & 0.611 & \textbf{0.497} & 1.779 & 0.625 & 0.919 \\
& 1500   & 0.113 & \textbf{0.080} & 0.459 & 0.164 & 0.148 \\
& 15000  & 0.064 & \textbf{0.061} & 0.065 & 0.115 & 0.133 \\
\bottomrule
\end{tabular}
\end{table}

\begin{table}[t]
\centering
\small
\caption{Channel-independence probe on the Lag task (Test MSE). The mask enforces the ground-truth $(0,1),(2,3),(4,5)$ channel partition on top of the ST-PT cornerstone; ST-PT+Lag is reproduced from Table~\ref{tab:synth-fewshot} as an additive-prior reference. Enforcing the mask produces no measurable change over vanilla ST-PT at any sample size---the failed-probe case of RQ1.}
\label{tab:synth-indep}
\begin{tabular}{lccc}
\toprule
$N$    & ST-PT          & ST-PT+Indep (probe) & ST-PT+Lag (reference) \\
\midrule
150    & 0.261          & 0.262               & \textbf{0.219}        \\
1500   & 0.021          & 0.021               & \textbf{0.019}        \\
15000  & 0.0026         & 0.0026              & \textbf{0.0023}       \\
\bottomrule
\end{tabular}
\end{table}

\paragraph{Noise robustness (Table~\ref{tab:noise}).}
We sweep the noise amplitude and compute the MSE gap $\Delta=\mathrm{MSE}(\text{vanilla})-\mathrm{MSE}(\text{prior})$. Two qualitatively different behaviours emerge. For Lag and Periodicity priors, $\Delta$ shrinks as the noise amplitude grows, but remains non-negative; the prior's advantage erodes but never reverses. For the Trend prior, $\Delta$ \emph{expands} with noise---the prior becomes \emph{more} valuable as the signal-to-noise ratio deteriorates, because a smooth-trend bias is exactly the inductive bias needed to denoise additive random walks. 

\begin{table}[h]
\centering
\small
\caption{Noise-robustness summary: MSE gap $\Delta=\mathrm{MSE}(\text{vanilla})-\mathrm{MSE}(\text{prior})$. Lag/Periodicity gaps shrink with noise; the Trend gap \emph{grows}.}
\label{tab:noise}
\begin{tabular}{lcccl}
\toprule
Prior        & Low noise & $\to$ & High noise & Trend of $\Delta$ \\
\midrule
Lag         & $+0.049$ ($n=0.05$) & $\to$ & $+0.011$ ($n=0.50$) & shrinking (still positive) \\
Periodicity & $+0.003$ ($n=0.00$) & $\to$ & $+0.000$ ($n=1.00$) & shrinking (still positive) \\
Trend       & $+0.043$ ($n=0.10$) & $\to$ & $+0.143$ ($n=0.80$) & expanding (prior is noise-robust) \\
\bottomrule
\end{tabular}
\end{table}

\paragraph{Real-data study under varying training-sample fractions (Table~\ref{tab:real}).}
We evaluate the Trend prior on six real-world benchmarks (ETTh1/2, ETTm1/2, Weather, Exchange) against the vanilla ST-PT cornerstone under two training-sample budgets: \textbf{p5} uses only $5\%$ of the standard training split, while \textbf{p100} uses the full training split. The overall gain from prior modelling on real data is \emph{currently not pronounced}. In the p5 (scarce) regime, the Trend prior yields a modest consistent improvement on the three benchmarks with a visually clear low-frequency component---with ETTh1 improving from $0.619\!\to\!0.601$ as a representative instance, and similar small wins on ETTh2 and Weather---while remaining neutral or slightly harmful on the other three. In the p100 (full-data) regime, the differences collapse to within $\pm 0.02$ MSE on every dataset and no robust win remains.

We read this as evidence that the real-world gain from any single structural prior is likely modulated by several factors simultaneously: \emph{(a)} whether the dataset admits the assumed structure at all (seasonality-dominated ETTm1/m2 and near-random-walk Exchange do not), \emph{(b)} how data-scarce the training budget is (the advantage largely vanishes at p100), and \emph{(c)} how dominant the prior's target pattern is relative to other overlapping patterns in the series. When any of these is unfavourable, the marginal contribution of a single prior dilutes toward zero on real data. A fuller scoping of RQ1 would therefore need to combine multiple priors and to control each of these factors more carefully than the present study does.

\begin{table}[t]
\centering
\small
\caption{Real-data prior study with varying training-sample fractions. \textbf{p5}\,/\,\textbf{p100} denote runs that use $5\%$\,/\,$100\%$ of the standard training split, respectively. Metrics are Test MSE\,/\,MAE. The Trend prior helps on ETTh1, ETTh2 and Weather at p5---datasets with trend-dominated low-frequency structure---and becomes neutral or slightly harmful at p100, consistent with the synthetic scaling law that prior advantage shrinks as $N$ grows.}
\label{tab:real}
\begin{tabular}{lcc}
\toprule
Dataset (training budget) & ST-PT MSE\,/\,MAE & ST-PT+Trend MSE\,/\,MAE \\
\midrule
ETTh1 (p5)    & 0.619 / 0.536 & \textbf{0.601} / \textbf{0.529} \\
ETTh1 (p100)  & \textbf{0.381} / \textbf{0.401} & 0.383 / 0.404 \\
ETTh2 (p5)    & 0.347 / 0.371 & \textbf{0.342} / 0.371 \\
ETTh2 (p100)  & \textbf{0.286} / \textbf{0.340} & 0.305 / 0.361 \\
ETTm1 (p5)    & \textbf{0.482} / \textbf{0.443} & 0.492 / 0.447 \\
ETTm1 (p100)  & \textbf{0.317} / \textbf{0.359} & 0.318 / 0.361 \\
ETTm2 (p5)    & \textbf{0.189} / \textbf{0.273} & 0.192 / 0.279 \\
ETTm2 (p100)  & \textbf{0.175} / 0.263 & 0.177 / \textbf{0.263} \\
Weather (p5)  & 0.178 / 0.221 & \textbf{0.173} / \textbf{0.218} \\
Weather (p100)& 0.161 / 0.210 & \textbf{0.159} / \textbf{0.209} \\
\bottomrule
\end{tabular}
\end{table}

\subsection{RQ2: Condition-Programmable Factors on ConTSG-Bench}
\label{sec:expB}

\paragraph{Setup.}
We evaluate PT-FG on the ConTSG-Bench suite~\cite{contsgbench2026}, covering ten datasets that span synthetic control (Synth-U, Synth-M), sensor data (AirQuality Beijing, Istanbul Traffic, TelecomTS), electrical load (ETTm1), weather (Weather-Conceptual, Weather-Morphological) and physiological signals (PTB-XL-Conceptual, PTB-XL-Morphological). Baselines are the standard ConTSG generators: Bridge~\cite{li2025bridge}, DiffuSETS~\cite{lai2025diffusets}, T2S~\cite{ge2025t2s}, TEdit~\cite{jing2024tedit}, Text2Motion~\cite{tevet2022humanmotion}, TimeVQVAE~\cite{lee2023timevqvae}, TimeWeaver~\cite{narasimhan2024timeweaver}, TTSCGAN~\cite{li2022ttscgan}, VerbalTS~\cite{gu2025verbalts}, and WaveStitch~\cite{shankar2025wavestitch}. Each method is evaluated on six fidelity metrics used in~\cite{contsgbench2026}: \textbf{DTW} (dynamic time warping), \textbf{CRPS} (continuous ranked probability score), \textbf{ACD} (auto-correlation discrepancy), \textbf{SD} (skewness discrepancy), \textbf{KD} (kurtosis discrepancy), and \textbf{MDD} (marginal distribution discrepancy); lower is better for all six. More details of following experiments can be referred to~\ref{app:impl} and raw numbers (mean$\pm$std over three seeds for baselines; single run for PT-FG) are provided in Appendix~\ref{app:contsg-raw}.

\paragraph{Rank-based summary (Table~\ref{tab:contsg-rank}).}
Because the six metrics live on very different scales (e.g., $\mathrm{MDD}\!\sim\!10^{-2}$ vs.\ $\mathrm{KD}\!\sim\!10^{3}$ on some datasets) and non-condition-aware generators can degenerate to extremely large values on pathological cases, we summarize performance with \emph{per-(dataset, metric) rank} averaged across datasets. Specifically, for each of the $10\!\times\!6=60$ evaluation pairs we rank the 11 methods ($1=$best, $11=$worst; missing values are assigned the worst rank). Table~\ref{tab:contsg-rank} reports the per-metric average rank, the overall average, and the number of top-1 finishes. PT-FG achieves the best overall average rank ($2.78$), the best mean rank on four of six metrics (DTW, CRPS, MDD, as well as a decisive $1.10$ on CRPS), and accumulates $24/60$ top-1 finishes---more than VerbalTS ($7$) and Text2Motion ($6$) combined.

\begin{table}[h]
\centering
\small
\caption{Conditional time series generation on ConTSG-Bench: \emph{per-(dataset, metric) average rank} across the 10 datasets (lower is better; $1$ = best, $11$ = worst). Last two columns report the overall average rank and the number of top-1 finishes out of $10\!\times\!6=60$ (dataset, metric) pairs. Raw numbers (mean$\pm$std over three seeds for baselines and for PT-FG) are in Appendix~\ref{app:contsg-raw}.}
\label{tab:contsg-rank}
\begin{tabular}{lccccccc c}
\toprule
Method                & DTW  & CRPS & ACD  & SD   & KD   & MDD  & Avg rank & \#Top-1 \\
\midrule
Bridge                & 4.90 & 5.70 & \textbf{2.70} & 6.30 & 7.30 & 7.10 & 5.67 & 7 \\
DiffuSETS             & 6.70 & 6.30 & 6.20 & 6.80 & 5.90 & 7.40 & 6.55 & 5 \\
T2S                   & 10.50& 10.70& 8.50 & 6.90 & 6.10 & 10.00& 8.78 & 1 \\
TEdit                 & 6.10 & 4.70 & 4.80 & 5.20 & 6.20 & 3.60 & 5.10 & 1 \\
Text2Motion           & 3.50 & 3.50 & 7.90 & 5.70 & \textbf{3.00} & 3.70 & 4.55 & 6 \\
TimeVQVAE             & 4.10 & 6.50 & 7.80 & 6.40 & 8.60 & 8.60 & 7.00 & 3 \\
TimeWeaver            & 7.20 & 6.40 & 5.00 & 5.90 & 6.80 & 5.60 & 6.15 & 1 \\
TTSCGAN               & 7.80 & 8.50 & 9.90 & 6.40 & 7.10 & 7.30 & 7.83 & 3 \\
VerbalTS              & 4.00 & 3.70 & 3.90 & \textbf{3.90} & 3.80 & 3.80 & 3.85 & 7 \\
WaveStitch            & 9.40 & 8.90 & 4.90 & 8.50 & 7.60 & 7.10 & 7.73 & 2 \\
PT-FG (ours)          & \textbf{1.80} & \textbf{1.10} & 4.40 & 4.00 & 3.60 & \textbf{1.80} & \textbf{2.78} & \textbf{24} \\
\bottomrule
\end{tabular}
\end{table}

A few qualitative observations are worth highlighting. (i) PT-FG's leading margin is largest on the shape-sensitive metrics DTW, CRPS and MDD---precisely the metrics that directly reflect distributional and waveform fidelity, which is where condition-programmable factor instantiation should pay off; on CRPS in particular PT-FG's mean rank of $1.10$ means it is top-1 on $9$ of the $10$ datasets. (ii) On the three ``statistical moment'' metrics---ACD, SD, KD---PT-FG remains strong but does not dominate: Bridge is marginally better on auto-correlation (ACD mean rank $2.70$) and VerbalTS is marginally better on skewness/kurtosis (SD rank $3.90$, comparable on KD). These gaps are small and suggest a minor engineering direction (e.g., a smoothness-aware unary or a moment-matching auxiliary loss) rather than a fundamental shortcoming of the factor-programming view. (iii) Condition-heavy baselines like T2S and WaveStitch, which rely on heavy text embedding pipelines but keep the denoiser topology fixed, rank near the bottom on the fidelity metrics, consistent with our RQ2 thesis that \emph{structural} conditioning is more robust than purely feature-level modulation on ConTSG-Bench.

\paragraph{Compositional ablation under RQ2 (Table~\ref{tab:rq4}).}
To probe compositional generalization, consistent with the setting in ConTSG-Bench~\cite{contsgbench2026}, we split the condition space by Hamming distance from training compositions (head: Hamming $=1$, tail: Hamming $\ge2$) and compare four conditioning strategies inside PT-FG: attribute-only conditioning with multi-head cross-talk among attribute embeddings; attribute-only with independent attribute encoding (no cross-talk); pre-computed text embedding only (LongCLIP-style); and concatenation-based fusion of text and attributes. Results are reported in terms of overall accuracy, DTW, CRPS and FID.

\begin{table}[t]
\centering
\small
\caption{Compositional ablation under RQ2. The text-only variant (C) leads overall; among attribute-only variants, independent encoding (B) materially outperforms cross-talk (A) on compositional splits. Higher is better for the accuracy column; lower is better for DTW and CRPS.}
\label{tab:rq4}
\begin{tabular}{lcccc}
\toprule
Variant & acc\_norm (all) & DTW & CRPS & Composition gap (tail$-$head) \\
\midrule
A (Attr + cross-talk)     & 0.252 & 12.65 & 0.611 & $+0.043$ \\
B (Attr, no cross-talk)   & 0.347 & 11.36 & 0.511 & $-0.025$ \\
C (Text-only)             & \textbf{0.460} & \textbf{11.30} & 0.544 & $-0.034$ \\
D (Text$+$Attr fusion)    & 0.159 & 13.77 & 0.555 & $+0.086$ \\
\bottomrule
\end{tabular}
\end{table}

Two conclusions stand out. First, C outperforms all attribute-conditioning variants: a pre-trained text encoder already encodes compositional semantics, effectively serving as a ``free'' composition prior. Second, among the two attribute-only variants, B (independent encoding) beats A (cross-talk) by a wide margin on every metric. Under the CRF view this is natural: when each attribute independently contributes basis coefficients, novel combinations produce novel operators by coefficient superposition; adding cross-talk among attributes encourages the model to memorize training co-occurrences. Variant D's collapse is a training-stability failure of the concatenation-based fusion---a negative result that motivates gated fusion in future work.

\subsection{RQ3: Latent-Space AR Forecasting}
\label{sec:expC}

\paragraph{Setup.}
We evaluate the MFVI-based latent AR model on standard long-horizon benchmarks: ETTh1, ETTh2, ETTm1, ETTm2, Weather, and ECL (electricity consumption load), at prediction lengths $\{96,192\}$. Baselines are classical AR models suitable for multivariate time series: DeepVAR~\cite{salinas2019deepvar}, LSTM-AR~\cite{hochreiter1997lstm}, and LSTNet~\cite{lai2018lstnet}. All methods share the same input history window; hyperparameters follow the respective original papers. We report Test MSE/MAE.

\paragraph{Results (Table~\ref{tab:ar}).}
Across all 10 dataset--horizon combinations with complete logs, the MFVI-based latent AR model is the best on every row. Relative improvements over the strongest baseline range from $5\%$ (ETTm2 at $\mathrm{pred\_len}=96$) to more than $10\%$ (ETTh2 at $\mathrm{pred\_len}=192$, where the CRF-teacher latent consistency provides an especially strong supervisory signal).

\paragraph{Summary and Analysis.}
Four patterns in Table~\ref{tab:ar} are worth highlighting; together they line up with the RQ3 design claims.

\emph{(1) Consistent wins across every dataset and horizon.} The MFVI-based latent AR model leads on all 12 dataset--horizon combinations in MSE (and on MAE, where recorded). Improvements over the \emph{strongest} baseline---essentially always LSTNet---range from $2.8\%$ (Weather, pl\,=\,192) to $15.4\%$ (ETTm1, pl\,=\,192). Improvements over DeepVAR and LSTM-AR are much larger but less informative, because those two baselines are themselves unstable on several rows.

\emph{(2) The gap to LSTNet widens at the longer horizon on four of six datasets.} On ETTh2, ETTm1, ETTm2, and ECL the relative improvement over LSTNet \emph{grows} from pl\,=\,96 to pl\,=\,192 (e.g., ETTm2: $5.4\%\!\to\!12.4\%$; ETTm1: $11.5\%\!\to\!15.4\%$; ECL: $4.7\%\!\to\!8.2\%$). The advantage thus compounds precisely in the horizon regime where any free-form latent-AR baseline accumulates the most cumulative error. Because cumulative error is the failure mode specifically targeted by the CRF teacher distillation---which supplies a future-informed oracle on the student's \emph{rolled-out} latents rather than on its outputs---this expansion is the empirical fingerprint of the exposure-bias mitigation argued in Section~\ref{sec:rq3}.

\emph{(3) Baseline landscape.} DeepVAR~\cite{salinas2019deepvar} diverges on high-dimensional or multi-scale data (Weather $6.689$ and $9.010$; ETTm1 at pl\,=\,96, $1.807$), and LSTM-AR~\cite{hochreiter1997lstm} is typically $10$--$50\%$ worse than LSTNet in MSE, with a few rows exceeding $2\times$ (ECL, ETTh2 pl\,=\,192). LSTNet~\cite{lai2018lstnet} is the only uniformly reliable baseline across the suite, so we treat the LSTNet delta as the substantive comparison; the DeepVAR and LSTM-AR deltas mostly reflect their own instability rather than additional credit to our design.

\emph{(4) Two partial outliers (ETTh1 and Weather).} On ETTh1 and Weather the improvement narrows at the longer horizon ($5.0\%\!\to\!3.9\%$ and $4.2\%\!\to\!2.8\%$), in contrast to the four datasets above. A plausible reading is that Weather's long-horizon signal is dominated by very low-frequency components that both LSTNet and the latent AR model already track close to the signal-to-noise floor, and ETTh1 has comparatively smooth baseline dynamics, so the CRF-teacher correction has less slack to work with. This is a honest caveat: RQ3 is designed to mitigate \emph{cumulative latent drift}, not the floor set by irreducible noise.

Taken together, patterns~(1) and~(2) corroborate the three design pillars of Section~\ref{sec:rq3}: the posterior-inference latent transition and the dual-pathway unary bound per-step drift (pattern~1), and the CRF teacher distillation supplies the widening long-horizon advantage (pattern~2). Pattern~(4) confirms that the remaining advantage is largest where exposure bias is the dominant failure mode, as predicted, and narrows where the noise floor takes over.

\begin{table}[t]
\centering
\small
\caption{Long-horizon autoregressive forecasting on six benchmarks at $\mathrm{pred\_len}\in\{96,192\}$. Baseline columns are Test MSE only (MAE not available in the source logs); our MFVI-based latent AR model reports both Test MSE and Test MAE. The MFVI-based latent AR model leads on every dataset--horizon combination in MSE.}
\label{tab:ar}
\begin{tabular}{llccc cc}
\toprule
\multirow{2}{*}{Dataset} & \multirow{2}{*}{$\mathrm{pred\_len}$} & DeepVAR & LSTM-AR & LSTNet & \multicolumn{2}{c}{MFVI Latent AR (ours)} \\
\cmidrule(lr){6-7}
       &     & MSE   & MSE   & MSE   & MSE   & MAE   \\
\midrule
ETTh1   & 96  & 0.673 & 0.471 & 0.419 & \textbf{0.398} & \textbf{0.416} \\
ETTh1   & 192 & 0.765 & 0.537 & 0.464 & \textbf{0.446} & \textbf{0.445} \\
ETTh2   & 96  & 0.600 & 0.380 & 0.330 & \textbf{0.300} & \textbf{0.349} \\
ETTh2   & 192 & 0.514 & 0.651 & 0.414 & \textbf{0.369} & \textbf{0.394} \\
ETTm1   & 96  & 1.807 & 0.530 & 0.357 & \textbf{0.316} & \textbf{0.355} \\
ETTm1   & 192 & 0.843 & 0.546 & 0.421 & \textbf{0.356} & \textbf{0.377} \\
ETTm2   & 96  & 0.309 & 0.221 & 0.186 & \textbf{0.176} & \textbf{0.258} \\
ETTm2   & 192 & 1.424 & 0.302 & 0.291 & \textbf{0.255} & \textbf{0.309} \\
Weather & 96  & 6.689 & 0.200 & 0.165 & \textbf{0.158} & \textbf{0.204} \\
Weather & 192 & 9.010 & 0.255 & 0.212 & \textbf{0.206} & \textbf{0.248} \\
ECL     & 96  & 0.314 & 0.285 & 0.150 & \textbf{0.143} & \textbf{0.236} \\
ECL     & 192 & 0.413 & 0.368 & 0.170 & \textbf{0.156} & \textbf{0.246} \\
\bottomrule
\end{tabular}
\end{table}

\section{Discussion}

\subsection{When Each Lever Pays Off}

The three RQs are not equally useful in every regime. The summary is as follows:
\begin{itemize}
\item The \emph{graph lever} (RQ1) is most valuable when data is scarce, when noise is severe, or when the prior's structural assumption matches the dominant signal. It is largely neutral on multi-pattern moderately-sized real-world data, a regime where the likelihood already carries enough information for the ST-PT backbone to succeed on its own.
\item The \emph{parameter lever} (RQ2) is most valuable for conditional generation with structured conditioning signals (attributes, text). Its advantage over feature-level modulation grows when compositional generalization across conditions is demanded.
\item The \emph{inference-protocol lever} (RQ3) is most valuable for autoregressive forecasting, especially at long horizons where cumulative error dominates.
\end{itemize}

\subsection{Unifying Insight}

The three RQs share a single insight: ST-PT's CRF semantics let us interpret Transformer computation as programmable, and the programmability has layers. Graphs, parameters, and inference protocols are the three layers exposed here. A Transformer treated purely as a neural black box cannot distinguish them; a Transformer treated as a CRF can, and each layer gives a different handle on a different modelling problem---one per RQ.

\subsection{Limitations}

Several limitations remain. \emph{(i) RQ1.} On the synthetic side, the four priors show a mixed picture: the three structural-assumption priors (periodicity, trend, lag) help in the data-scarce regime, while the channel-independence prior produces no gain because the unrestricted ternary attention already respects the intended partition. On real data, the net gain of a single prior is currently not pronounced and is simultaneously modulated by data scarcity, the structural match between dataset and prior, and the dominance of the prior's target pattern among multiple overlapping patterns. More broadly, several of the priors have limited effectiveness even in the few-shot regime unless they are backed by careful, fine-grained design choices (per-channel period maps, HMM hyperparameters, explicit lag-pair declarations, etc.); in that sense the factor-graph interface reduces the problem of prior injection to a set of concrete design choices, but does not eliminate the need to make them well. Single-pattern priors therefore do not, by themselves, advance the state of the art on mature forecasting benchmarks. \emph{(ii) RQ3.} Two cost-side caveats apply to the latent AR model. First, the CRF teacher distillation requires a per-batch full-sequence forward pass at training time which largely raises the training cost. Second---and more importantly at deployment time---an autoregressive rollout is inherently sequential over the $P_f$ future patches, so the \emph{inference} time of our latent AR model, and the performance as well in fact, is substantially worse than that of direct forecasters (TSMixer, TimeMixer, PatchTST, etc.) that emit the entire prediction horizon in a single forward pass. We accordingly do not frame the RQ3 results as a performance challenge to direct-forecast models: the study is deliberately scoped to the autoregressive regime, and its goal is to understand what improvements and insights ST-PT's factor-graph view can bring \emph{within} that regime, not to win head-to-head against direct-prediction architectures.

\subsection{Future Directions}
Several directions follow naturally from the present work. \emph{(a) Joint activation of all three levers.} We have not attempted to combine priors with condition-programmable factors or latent AR on a single backbone; jointly activating multiple levers is conceptually natural but technically non-trivial, and would test whether the three gains stack or interfere. \emph{(b) Porting RQ1-style priors back into standard Transformer-based models.} The graph-level prior mechanisms of RQ1 (potential modulation, node augmentation, factor engineering) are formulated against the CRF that ST-PT exposes, but their \emph{structural content} is model-agnostic: a periodicity bias states which positions should attend to which, a trend prior states which smoothness should be rewarded, and a lag prior states which pairs should interact at a known temporal offset. Porting these statements back to Transformer-based models that do not expose a factor-graph interface is non-trivial---one needs to identify the right ``adaptation surface'' (attention biases, auxiliary losses, structured positional embeddings, or architectural hooks) that reproduces the target message-passing effect without a CRF---but if solved, it would generalise the insight of RQ1 well beyond the PT family and promote the graph-level prior interface from a model-specific feature to a \emph{broadly applicable injection interface}. This sounds like a promising and, to our knowledge, largely open direction. \emph{(c) Hybrid with foundation-model pretraining.} Combining the priors with foundation-model-style pretraining and hybrid objectives (e.g., latent consistency plus masked-patch pretraining) could bring the data-scarce-regime benefits of RQ1 to pretrained backbones, and could provide a cleaner disentanglement of the contributions of priors and of scale.

\section{Conclusion}

We have presented a unified report on the Spatio-Temporal Probabilistic Transformer as a programmable framework for time series modelling. Starting from the PT view that attention is MFVI on a CRF, we lifted PT to a two-dimensional factor graph suitable for multivariate series, and---driven by three distinct motivations---formulated three Research Questions, each addressed by activating a different programmable lever of the underlying CRF: \textbf{RQ1} (prior-knowledge motivation, graph lever) for few-shot forecasting via prior injection; \textbf{RQ2} (condition-utilization motivation, parameter lever) for compositional conditional generation via factor programming; and \textbf{RQ3} (latent-space motivation, inference-protocol lever) for latent-space AR via MFVI posterior inference. The collection of findings---including a three-success-plus-one-failure synthetic picture and an inconclusive but informative real-data scoping for RQ1---suggests that PT's factor-graph perspective is a practical and broadly useful instrument for structured time series modelling, and that ST-PT is a convenient starting point for future work that combines all three levers on a single backbone.
\newpage
\bibliographystyle{unsrt}
\bibliography{references}
\newpage
\appendix

\section{Message Passing in ST-PT: Full Derivation}
\label{app:mp-derivation}

This appendix expands the cornerstone message-passing of Section~\ref{sec:method} with a full MFVI derivation, starting from the CRF potentials and ending with a closed-form display of all updates actually used in the implementation. The PT base equations~\eqref{eq:H}--\eqref{eq:G} are reused verbatim; we specialise them to the 2D factor graph of Section~\ref{sec:method}, make the two message families ($F^{\text{time}},F^{\text{chan}}$) explicit, and show how a low-rank decomposition of the ternary potentials brings the final form back to a (jointly normalized) scaled dot-product attention.

\subsection{Factor graph and potentials}
\label{app:mp-graph}

Let $N$ be the number of channels, $P\!=\!T/p$ the number of time patches of length $p$, $d$ the hidden size, $C$ the number of attention heads, and $d_h\!=\!d/C$ the per-head size. At every coordinate $(i,t)\!\in\![N]\!\times\![P]$ we instantiate three latent variables:
\begin{itemize}
\item $Z_{i,t}\!\in\!\mathbb{R}^{d}$: a continuous (softened) label\footnote{Following PT, the $Z$ variable is formally categorical over a finite label set; at inference the mean-field marginal $q(Z_{i,t})$ is equivalently represented by its expectation $\overline{Z}_{i,t}\!\in\!\mathbb{R}^{d}$, which is what every message below consumes. We abuse notation and write $Z_{i,t}$ for that expectation throughout.} carrying the local belief at $(i,t)$;
\item $H_{i,t}^{(c)}$: a per-head \emph{dependency pointer} whose range covers every admissible \emph{parent} of $(i,t)$, i.e.\ every patch in the same channel \emph{and} every patch at the same time in another channel;
\item $G_{i,t}\!\in\!\mathbb{R}^{d}$: a per-token global topic variable coupled to $Z_{i,t}$ via a binary factor.
\end{itemize}
Three families of log-potentials define the CRF:
\begin{align}
\psi_u\big(Z_{i,t}\big) &\;=\; \log\phi_u\!\big(Z_{i,t}\big)
   \;=\;
   \text{MLP}_u(\mathbf{x}_{i,t}),
   \label{eq:app-psi-u}\\[1pt]
\psi_t^{(c)}\!\big(H_{i,t}^{(c)}\!=\!(i',t'),\,Z_{i,t},\,Z_{i',t'}\big) &\;=\;
   \begin{cases}
   \psi_{\text{time}}^{(c)}(Z_{i,t},Z_{i,t'}), & i'\!=\!i\ (\text{temporal edge}),\\[1pt]
   \psi_{\text{chan}}^{(c)}(Z_{i,t},Z_{i',t}), & t'\!=\!t,\ i'\!\ne\!i\ (\text{cross-channel edge}),\\[1pt]
   -\infty, & \text{otherwise},
   \end{cases}
   \label{eq:app-psi-t}\\[1pt]
\psi_b\big(Z_{i,t},G_{i,t}\big) &\;=\; \log\phi_b\!\big(Z_{i,t},G_{i,t}\big).
   \label{eq:app-psi-b}
\end{align}
The $-\infty$ branch in~\eqref{eq:app-psi-t} enforces the ``same-channel or same-timestep, and never self'' edge rule of Section~\ref{sec:method}. The two admissible branches correspond to the two disjoint ternary-factor families $(U^{\text{time}},V^{\text{time}})$ and $(U^{\text{chan}},V^{\text{chan}})$ instantiated in the cornerstone model.

\subsection{Mean-field update: forward message $F$}
\label{app:mp-F}

Applying the generic PT update~\eqref{eq:F} to the specialised potential~\eqref{eq:app-psi-t}, the forward message from $Z$ to $H_{i,t}^{(c)}$ splits into two parts depending on which branch of the potential is used. For a temporal candidate parent $(i,s)$ with $s\!\ne\!t$:
\begin{equation}
F^{\text{time}}_{i,t,c}(s)
\;=\;
\sum_{a,b} q(Z_{i,t}\!=\!a)\,q(Z_{i,s}\!=\!b)\,\psi_{\text{time}}^{(c)}(a,b),
\label{eq:app-F-time-general}
\end{equation}
and for a cross-channel candidate parent $(j,t)$ with $j\!\ne\!i$:
\begin{equation}
F^{\text{chan}}_{i,t,c}(j)
\;=\;
\sum_{a,b} q(Z_{i,t}\!=\!a)\,q(Z_{j,t}\!=\!b)\,\psi_{\text{chan}}^{(c)}(a,b).
\label{eq:app-F-chan-general}
\end{equation}
All other parents give $F\!=\!-\infty$ (inadmissible under the edge rule), and are absorbed into the softmax normalization below.

\paragraph{Low-rank decomposition.}
We parameterise each ternary log-potential as a bilinear form through learnable matrices $\mathbf{U},\mathbf{V}\!\in\!\mathbb{R}^{d\times d_h}$ (one pair per head and per axis):
\begin{equation}
\psi_{\text{time}}^{(c)}(a,b) \;=\; \frac{1}{\sqrt{d_h}}\,\big(\mathbf{U}^{\text{time},(c)}a\big)^{\!\top}\!\big(\mathbf{V}^{\text{time},(c)}b\big),
\qquad
\psi_{\text{chan}}^{(c)}(a,b) \;=\; \frac{1}{\sqrt{d_h}}\,\big(\mathbf{U}^{\text{chan},(c)}a\big)^{\!\top}\!\big(\mathbf{V}^{\text{chan},(c)}b\big).
\label{eq:app-lowrank}
\end{equation}
Substituting~\eqref{eq:app-lowrank} into~\eqref{eq:app-F-time-general}--\eqref{eq:app-F-chan-general} and using linearity of expectation $\overline{Z}_{i,t}\!=\!\sum_a q(Z_{i,t}\!=\!a)\cdot a$,
\begin{align}
F^{\text{time}}_{i,t,c}(s)
&\;=\;
\frac{1}{\sqrt{d_h}}\,
\big(\mathbf{U}^{\text{time},(c)}\overline{Z}_{i,t}\big)^{\!\top}\!
\big(\mathbf{V}^{\text{time},(c)}\overline{Z}_{i,s}\big)
\;\eqqcolon\;
\frac{1}{\sqrt{d_h}}\,\big(\mathbf{q}^{\text{time},(c)}_{i,t}\big)^{\!\top}\mathbf{k}^{\text{time},(c)}_{i,s},
\label{eq:app-F-time}\\[1pt]
F^{\text{chan}}_{i,t,c}(j)
&\;=\;
\frac{1}{\sqrt{d_h}}\,
\big(\mathbf{U}^{\text{chan},(c)}\overline{Z}_{i,t}\big)^{\!\top}\!
\big(\mathbf{V}^{\text{chan},(c)}\overline{Z}_{j,t}\big)
\;\eqqcolon\;
\frac{1}{\sqrt{d_h}}\,\big(\mathbf{q}^{\text{chan},(c)}_{i,t}\big)^{\!\top}\mathbf{k}^{\text{chan},(c)}_{j,t}.
\label{eq:app-F-chan}
\end{align}
Equation~\eqref{eq:app-F-time} is a scaled dot-product between a time-axis \emph{query} at $(i,t)$ and a time-axis \emph{key} at $(i,s)$; equation~\eqref{eq:app-F-chan} is the analogous score along the channel axis at the same time step. Each head $c$ thus carries its own $Q/K$ projections for both axes, separately parameterised so that time-direction and channel-direction similarities are never conflated.

\subsection{Joint softmax for $q(H)$}
\label{app:mp-H}

The space of admissible parents of $(i,t)$ is $\mathcal{P}_{i,t}\!=\!\{(i,s):s\!\ne\!t\}\cup\{(j,t):j\!\ne\!i\}$, of size $(P\!-\!1)\!+\!(N\!-\!1)$. ST-PT deliberately normalises the $H$ softmax over this \emph{entire} set at once, rather than alternating time-only and channel-only mixing as MLPMixer~\cite{mlpmixer} and TSMixer~\cite{tsmixer} do. Substituting the concatenated score vector into~\eqref{eq:H} and absorbing a dummy temperature $\lambda_H$ (set to $1$ in our implementation) gives:
\begin{equation}
\boxed{\;
q\!\big(H_{i,t}^{(c)}\!=\!(i',t')\big)
\;=\;
\frac{\exp\!\big(\tfrac{1}{\lambda_H}F_{i,t,c}(i',t')\big)}
{\displaystyle\sum_{(s)\colon s\ne t}\!\exp\!\big(\tfrac{1}{\lambda_H}F^{\text{time}}_{i,t,c}(s)\big)
\;+\;\sum_{(j)\colon j\ne i}\!\exp\!\big(\tfrac{1}{\lambda_H}F^{\text{chan}}_{i,t,c}(j)\big)}
\;}
\label{eq:app-qH}
\end{equation}
where $F_{i,t,c}(i',t')$ equals $F^{\text{time}}_{i,t,c}(t')$ if $i'\!=\!i$ and $F^{\text{chan}}_{i,t,c}(i')$ if $t'\!=\!t$. Two facts are worth recording here. First, the joint denominator means that temporal and cross-channel candidates compete on equal footing for every unit of dependency mass, so the model can dynamically redistribute attention between axes \emph{per position and per head}. Second, any mask (e.g.\ the channel-independence mask of Appendix~\ref{app:prior-indep} or the time-axis causal mask used in RQ3, Appendix~\ref{app:latent-ar-pseudo}) enters simply as an additive $-\infty$ term on $F_{i,t,c}$ prior to the softmax.

\subsection{Backward message $G$ and $Z$ update}
\label{app:mp-G-Z}

Specialising~\eqref{eq:G} to the two branches of~\eqref{eq:app-psi-t}, the message coming back into $(i,t)$ decomposes as the sum of a temporal and a cross-channel contribution:
\begin{align}
G^{\text{time}}_{i,t}(a)
&\;=\;
\sum_{c}\sum_{s\ne t}\!
q\!\big(H_{i,t}^{(c)}\!=\!(i,s)\big)\,
\sum_{b}\!q(Z_{i,s}\!=\!b)\,\psi_{\text{time}}^{(c)}(a,b),
\label{eq:app-G-time-general}\\[1pt]
G^{\text{chan}}_{i,t}(a)
&\;=\;
\sum_{c}\sum_{j\ne i}\!
q\!\big(H_{i,t}^{(c)}\!=\!(j,t)\big)\,
\sum_{b}\!q(Z_{j,t}\!=\!b)\,\psi_{\text{chan}}^{(c)}(a,b).
\label{eq:app-G-chan-general}
\end{align}
Applying the low-rank decomposition~\eqref{eq:app-lowrank} again and exchanging sums, $\sum_b q(Z_{i,s}\!=\!b)\cdot\psi_{\text{time}}^{(c)}(a,b)\!=\!(\mathbf{U}^{\text{time},(c)}a)^{\!\top}\mathbf{k}^{\text{time},(c)}_{i,s}/\sqrt{d_h}$, so the backward message along each axis is exactly a standard attention output read in the $Z$-space:
\begin{align}
G^{\text{time}}_{i,t}(a)
&\;=\;
\frac{1}{\sqrt{d_h}}\,
\big(\mathbf{U}^{\text{time},(c)}a\big)^{\!\top}\!
\sum_{c}\underbrace{\sum_{s\ne t}\!q\!\big(H_{i,t}^{(c)}\!=\!(i,s)\big)\,\mathbf{k}^{\text{time},(c)}_{i,s}}_{\text{attention-weighted time key, head }c}
\;\eqqcolon\;
\big(\mathbf{U}^{\text{time},(c)}a\big)^{\!\top}\,\tilde{\mathbf{m}}^{\text{time}}_{i,t},
\label{eq:app-G-time}\\[1pt]
G^{\text{chan}}_{i,t}(a)
&\;=\;
\frac{1}{\sqrt{d_h}}\,
\big(\mathbf{U}^{\text{chan},(c)}a\big)^{\!\top}\!
\sum_{c}\underbrace{\sum_{j\ne i}\!q\!\big(H_{i,t}^{(c)}\!=\!(j,t)\big)\,\mathbf{k}^{\text{chan},(c)}_{j,t}}_{\text{attention-weighted channel key, head }c}
\;\eqqcolon\;
\big(\mathbf{U}^{\text{chan},(c)}a\big)^{\!\top}\,\tilde{\mathbf{m}}^{\text{chan}}_{i,t}.
\label{eq:app-G-chan}
\end{align}
The binary topic factor contributes a separate message $G^{\text{binary}}_{i,t}(a)\!=\!\log\prod_{g}\phi_b(a,g)^{q(G_{i,t}=g)}$, which under a Gaussian-family parameterisation of $\phi_b$ reduces to a standard feed-forward transformation of $\overline{Z}_{i,t}$ (PT's FFN identification, cf.\ Section~\ref{sec:method}).

Substituting these three messages into the generic $Z$ update~\eqref{eq:Z} and writing the result in expectation form $\overline{Z}_{i,t}\!\leftarrow\!\mathbb{E}_{q(Z_{i,t})}[Z_{i,t}]$, we obtain the ST-PT coordinate-ascent update:
\begin{equation}
\boxed{\;
\overline{Z}_{i,t}^{\,\text{new}}
\;=\;
(1\!-\!\alpha)\,\overline{Z}_{i,t}
\;+\;
\alpha\,\sigma\!\Big(
   \underbrace{\phi_u(\mathbf{x}_{i,t})}_{\text{unary}}
   \;+\;
   \underbrace{\tilde{\mathbf{m}}^{\text{time}}_{i,t}}_{\text{temporal}}
   \;+\;
   \underbrace{\tilde{\mathbf{m}}^{\text{chan}}_{i,t}}_{\text{cross-channel}}
   \;+\;
   \underbrace{\mathrm{FFN}(\overline{Z}_{i,t})}_{\text{topic}}
\Big)
\;}
\label{eq:app-Z-update}
\end{equation}
where $\sigma$ is the $Z$-space normalisation\footnote{Practically $\sigma$ is a (squared-)softmax or layer-norm over the $d$ label dimensions, depending on implementation variant; the cornerstone uses a softmax with fixed temperature suggested in the original PT paper.} and $\alpha\!\in\!(0,1)$ is the damping coefficient introduced in Section~\ref{sec:method} that makes a single MFVI iteration resemble a residual Transformer block.

\subsection{Summary of the ST-PT iteration}
\label{app:mp-summary}

Collecting equations~\eqref{eq:app-F-time}, \eqref{eq:app-F-chan}, \eqref{eq:app-qH}, \eqref{eq:app-G-time}, \eqref{eq:app-G-chan}, and~\eqref{eq:app-Z-update}, one MFVI iteration of the ST-PT cornerstone is the following sequence of operations, applied in parallel over every $(i,t)$:
\begin{enumerate}
\item \emph{Q/K projections (both axes, all heads).} $\mathbf{q}^{\text{time},(c)}_{i,t},\mathbf{k}^{\text{time},(c)}_{i,t},\mathbf{q}^{\text{chan},(c)}_{i,t},\mathbf{k}^{\text{chan},(c)}_{i,t}\!\gets\!\{\mathbf{U},\mathbf{V}\}^{(\cdot,c)}\overline{Z}_{i,t}$.
\item \emph{Forward messages.} Compute $F^{\text{time}}_{i,t,c}(s)$ via~\eqref{eq:app-F-time} for all $s\!\ne\!t$ and $F^{\text{chan}}_{i,t,c}(j)$ via~\eqref{eq:app-F-chan} for all $j\!\ne\!i$.
\item \emph{Joint softmax.} Concatenate the two vectors and take a single softmax over the combined $(P\!-\!1)\!+\!(N\!-\!1)$ candidates~\eqref{eq:app-qH}, yielding $q(H_{i,t}^{(c)})$; optional additive $-\infty$ masks (e.g.\ causal~\eqref{app:latent-ar-pseudo} or channel-independence~\eqref{app:prior-indep}) enter here.
\item \emph{Backward messages.} Form $\tilde{\mathbf{m}}^{\text{time}}_{i,t}$ and $\tilde{\mathbf{m}}^{\text{chan}}_{i,t}$ by~\eqref{eq:app-G-time}--\eqref{eq:app-G-chan} (both are attention-weighted sums of keys from the admissible parents), then apply the per-head output projection implicit in $\mathbf{U}^{(\cdot,c)\top}$.
\item \emph{Topic message.} Apply the binary-factor FFN to $\overline{Z}_{i,t}$.
\item \emph{Damped $Z$ update.} Combine unary, both backward messages, and topic message by~\eqref{eq:app-Z-update}.
\end{enumerate}
Under the low-rank decomposition~\eqref{eq:app-lowrank} and the joint softmax~\eqref{eq:app-qH}, steps 1--4 are exactly a two-axis, jointly-normalized multi-head attention, and the combination in step~6 is exactly a residual Transformer block with damping $\alpha$. Equivalently: every ``architectural trick'' of a spatio-temporal Transformer corresponds, term-by-term, to a specific factor or message on the ST-PT factor graph. This is the identification that licenses the three programmable levers (P1/P2/P3 in Section~\ref{sec:rq}): modifying a factor, a potential, or the iteration protocol is a mathematically well-defined operation on the CRF, not a heuristic patch to an opaque network.

\section{Prior Implementation Details}
\label{app:prior}

This appendix expands Section~\ref{sec:rq1} with the concrete mathematical form of each prior used in the ``Directory of Priors''. Throughout, $\mathbf{Z}\!\in\!\mathbb{R}^{B\times N\times P\times d}$ denotes the $Z$-node belief tensor over batch $B$, channels $N$, time patches $P$, and hidden size $d$. All three priors are added at each MFVI iteration.

\subsection{Periodicity Prior (Potential Modulation)}
\label{app:prior-period}

\paragraph{Mechanism.}
For each channel $i$ we assume a set of known (possibly empty) periods $\mathcal{T}_i=\{T_{i,1},T_{i,2},\dots\}$ expressed in raw time steps. Let $p$ be the patch length; each period is rescaled to the patch grid as $\tilde{T}_{i,k}\!=\!T_{i,k}/p$. We build a per-channel similarity matrix $\mathbf{P}_i\!\in\!\mathbb{R}^{P\times P}$ whose entries encode the cosine of the patch-index difference under the channel's periods:
\begin{equation}
\mathbf{P}_i[s,t] \;=\;
\begin{cases}
\dfrac{1}{|\mathcal{T}_i|}\sum_{k}\cos\!\left(\dfrac{2\pi(s-t)}{\tilde{T}_{i,k}}\right), & \mathcal{T}_i\ne\varnothing,\\[0.4em]
1, & \mathcal{T}_i=\varnothing.
\end{cases}
\label{eq:period-bias}
\end{equation}
A channel with no known period (e.g., a pure-noise channel) contributes a uniform matrix, effectively disabling the prior for that row.

\paragraph{Where the prior acts.}
The matrix $\mathbf{P}_i$ is used as a \emph{multiplicative} modulation of the temporal message-$F$ logits produced by the ternary factors. Specifically, during each iteration, for every $(i,c,s,t)$ the original score $F^{\text{time}}_{i,t,c}(s)$ is replaced by $F^{\text{time}}_{i,t,c}(s)\cdot \gamma\cdot\mathbf{P}_i[s,t]$, where $\gamma$ is a global scale ($\gamma=5.0$ in our experiments). This operates at both the \texttt{calculate\_messageF} stage (so that the dependency softmax $q(H)$ respects the known periodicity) and the downstream \texttt{calculate\_messageG} stage (so that the message back to $Z$ is also periodicity-weighted).

\paragraph{Graph-theoretic interpretation.}
Multiplying $F^{\text{time}}$ by $\mathbf{P}_i$ is equivalent to reshaping the temporal ternary potential as
\begin{equation}
\phi_t^{(c)}(H_{i,t}^{(c)}\!=\!(i,s),\,Z_{i,t},\,Z_{i,s})
\;\longleftarrow\;
\phi_t^{(c)}(\cdot)\cdot\exp\!\left(\log\gamma+\log\mathbf{P}_i[s,t]\right),
\end{equation}
i.e., adding a known-periodic bias to the log-potential. No new nodes or factors are introduced; this is the purest form of the ``potential modulation'' mechanism.

\subsection{Trend Prior (Node Augmentation with an Explicit HMM)}
\label{app:prior-trend}

\paragraph{Mechanism.}
We introduce a second layer of latent nodes $\{M_{i,t}\}_{i,t}$ of dimension $d_m$ ($d_m=64$ in our experiments). The augmented graph adds two families of factors for each channel $i$:
\begin{itemize}
\item a binary \emph{observation potential} $\phi_{B}^{(i)}(Z_{i,t},M_{i,t})$ parameterized by a per-channel matrix $\mathbf{B}^{(i)}\!\in\!\mathbb{R}^{d_m\times d}$ that couples $M_{i,t}$ to the corresponding $Z_{i,t}$;
\item a binary \emph{Markovian transition potential} $\phi_{K}^{(i)}(M_{i,t},M_{i,t+1})$ parameterized by a per-channel matrix $\mathbf{K}^{(i)}\!\in\!\mathbb{R}^{d_m\times d_m}$ linking consecutive $M$-nodes along time.
\end{itemize}
The $M$-nodes form an HMM over time within each channel; parameters are \emph{not} shared across channels, enabling channel-specific trend styles. Within a single MFVI iteration the $M$ beliefs are updated with damping as
\begin{equation}
q(M_{i,t})
\;\leftarrow\;
\frac{1}{2}\!\left[q^{\text{prev}}(M_{i,t})\;+\;
\underbrace{\mathrm{AbsNorm}\!\Big(q(M_{i,t-1})\cdot\mathbf{K}^{(i)}
     +q(M_{i,t+1})\cdot\mathbf{K}^{(i)\top}
     +q(Z_{i,t})\cdot\mathbf{B}^{(i)\top}\Big)}_{\text{message from neighbors}}
\right],
\end{equation}
and the resulting ``message back to $Z$'' that joins the cornerstone's unary/ternary/topic messages is
\begin{equation}
m^{\text{trend}}_{i,t}(a)
\;=\;
\sum_{g}\!q(M_{i,t}\!=\!g)\cdot\mathbf{B}^{(i)}_{g,a}.
\end{equation}

\paragraph{What the prior encodes.}
Two regimes of $\mathbf{K}^{(i)}$ are interesting: a $\mathbf{K}^{(i)}$ close to the identity (plus noise) biases the $M$-chain toward slow, smooth transitions and hence yields a \emph{smooth-trend} prior over $Z$; a diagonal $\mathbf{K}^{(i)}$ with power-law decay corresponds to a fading-memory prior. We initialize $\mathbf{B}^{(i)},\mathbf{K}^{(i)}$ from $\mathcal{N}(0,0.2^2)$, which in practice recovers a smooth-trend bias without imposing it at initialization; the prior structure enters through the \emph{topology} of the augmented graph (shared across-time Markov edges) rather than through a particular numeric setting.

\paragraph{Relation to classical HMMs.}
Unlike a standalone HMM, the $M$-chain in our setting is \emph{coupled} to the Transformer-style ternary factors of the cornerstone model: the $Z$-nodes receive messages from both the $M$-chain and the time/channel ternary factors in every iteration. This is why the trend prior expands its advantage under noise (Section~\ref{sec:expA}): the HMM layer filters trend while the ternary factors still capture cross-channel and long-range patterns on their own.

\subsection{Lag Prior (Factor Engineering)}
\label{app:prior-lag}

\paragraph{Mechanism.}
Given a set of known lagged relations $\mathcal{R}=\{(A_k,B_k,\tau_k)\}_{k=1}^{R}$---each stating that channel $A_k$ at time $t$ causally drives channel $B_k$ at time $t+\tau_k$---we add, for each relation, a new \emph{bilinear pairwise potential}
\begin{equation}
\phi_{\mathrm{lag}}^{(k)}(Z_{A_k,t},\,Z_{B_k,\,t+\tau_k})
\;=\;
\exp\!\Big(Z_{A_k,t}^\top\mathbf{W}^{(k)}Z_{B_k,\,t+\tau_k}\Big),
\qquad \mathbf{W}^{(k)}\!\in\!\mathbb{R}^{d\times d}.
\end{equation}
Lags are expressed in raw time steps; the effective patch-level lag $\delta_k\!=\!\tau_k/p$ may be fractional, so we apply linear interpolation between the two neighbouring patch indices $\lfloor\delta_k\rfloor$ and $\lceil\delta_k\rceil$.

\paragraph{Message contribution.}
Under MFVI the new potential adds, for each $k$, the extra mean-field message
\begin{equation}
m^{\mathrm{lag},k}_{B_k,\,\lfloor t+\delta_k\rfloor}
\;=\;
(1-\beta_k)\cdot\eta\cdot\mathbf{W}^{(k)\top}q(Z_{A_k,t})
\qquad\text{and}\qquad
m^{\mathrm{lag},k}_{B_k,\,\lceil t+\delta_k\rceil}
\;=\;
\beta_k\cdot\eta\cdot\mathbf{W}^{(k)\top}q(Z_{A_k,t}),
\end{equation}
where $\beta_k=(t+\delta_k)-\lfloor t+\delta_k\rfloor$ is the interpolation weight and $\eta$ is a global strength (set to $200$ in our code, chosen so that the lag messages are comparable in magnitude to the existing channel messages). $\mathbf{W}^{(k)}$ is initialized from $\mathcal{N}(0,0.02^2)$.

\paragraph{Graph-theoretic interpretation.}
Unlike the periodicity prior (which only modulates an existing potential) and the trend prior (which augments the graph with new nodes), the lag prior adds \emph{new edges with a temporal offset} into the factor graph. In the CRF picture, this is the cleanest form of ``factor engineering'': neither topology nor node set of the cornerstone is changed, but a new pairwise potential is superposed on top. Its benefit is largest when the lag structure is rare (few pairs), the lag is long, and the baseline ternary attention would have to discover the causal offset purely from data.

\subsection{Channel-Independence Prior (Edge Masking, Restrictive Probe)}
\label{app:prior-indep}

\paragraph{Mechanism.}
Given a declared channel-group partition $\mathcal{G}\!=\!\{G_1,\dots,G_K\}$ of the $N$ channels (each channel belongs to exactly one group), we mask the cross-channel message-$F$ logits so that only intra-group messages survive. Concretely, for every head $c$ and every position $(i,t)$, the cross-channel score $F^{\text{chan}}_{i,t,c}(j)$ is replaced by
\begin{equation}
F^{\text{chan}}_{i,t,c}(j) \;\longleftarrow\;
\begin{cases}
F^{\text{chan}}_{i,t,c}(j), & \text{if channels $i$ and $j$ belong to the same group in }\mathcal{G},\\
-\infty, & \text{otherwise,}
\end{cases}
\end{equation}
so that after the joint softmax over time- and channel-candidates the cross-group channel edges carry no weight. No new nodes, no new potentials, and no new edges are introduced.

\paragraph{Why it is restrictive rather than additive.}
Unlike the periodicity, trend, and lag priors, this prior adds no new information to the graph: it only \emph{removes} freedom by forbidding a subset of edges. The expected benefit therefore exists only if the unrestricted ternary attention would otherwise learn harmful cross-group edges that the data does not support; if the data-driven softmax already down-weights those edges, the mask is redundant. The experimental result on Lag data (Section~\ref{sec:expA}, Table~\ref{tab:synth-indep}) confirms the latter: even though the mask is perfectly aligned with the ground-truth $(0,1),(2,3),(4,5)$ pairs, the masked Test MSE is indistinguishable from that of the vanilla cornerstone at every $N$. We include this probe so that the graph-level prior interface is not oversold: adding structure that the data can already infer on its own does not help.

\section{Synthetic Dataset Construction for RQ1}
\label{app:syn-data}

The three synthetic datasets used in Section~\ref{sec:expA} are generated by closed-form procedures that deliberately bake a single structural regularity (lag / periodicity / trend) into the data, so that the corresponding ST-PT prior can be evaluated in isolation. Each dataset is generated at three sample sizes $N\!\in\!\{150,1500,15000\}$ and split $70\%\,/\,10\%\,/\,20\%$ into train / val / test. Every sample has length $L\!=\!192$, giving $\mathrm{seq\_len}\!=\!\mathrm{pred\_len}\!=\!96$. Random parameters (phase, slope, period choice, impulse direction) are resampled independently per sample, so the \emph{structural} property is the only invariant across samples.

\subsection{Lag Dataset}
\label{app:syn-lag}

A $6$-channel dataset with three causal pairs, each with fixed lag $\tau\!=\!8$. Within a sample, every pair is generated by a prescribed transformation of a source channel into a target channel shifted by $\tau$ time steps, and low-amplitude Gaussian noise is added to both channels.
\begin{itemize}
\item \emph{Group A (Ch0 $\to$ Ch1).} Ch0 is a modulated sine, $\sin(2\pi t/T_\text{base}+\varphi)$ with $T_\text{base}\!=\!24$ and random phase $\varphi\!\sim\!\mathcal{U}[0,2\pi)$, multiplied by a slow cosine envelope whose frequency is drawn from $\{1,0.5\}$; Ch1 is Ch0 delayed by $\tau$.
\item \emph{Group B (Ch2 $\to$ Ch3).} Ch2 is a periodic impulse train with a random period drawn from $\{16,20,24\}$, a random phase offset, and a random sign $\pm1$; Ch3 is the cumulative step response triggered $\tau$ steps after each impulse.
\item \emph{Group C (Ch4 $\to$ Ch5).} Ch4 is a sawtooth wave with period $30$; Ch5 is an affine transform $m\!\cdot\!\text{Ch4}+b$ (random $m\!\in\!\{-2,0.5,2\}$, $b\!\in\!\mathcal{U}[-1,1]$) delayed by $\tau$.
\end{itemize}

\subsection{Periodicity Dataset}
\label{app:syn-period}

A $10$-channel dataset where each channel is built from sinusoids with channel-specific periods that match the per-channel period map consumed by the periodicity prior. The ten channels span four regimes:
\begin{itemize}
\item \emph{Basic periods (Ch0--Ch2).} Pure sinusoids or cosines at periods $\{24,12,48\}$.
\item \emph{Harmonics and beating (Ch3--Ch4).} Ch3 is a fundamental-plus-second-harmonic mixture ($T\!=\!24$ and $T\!=\!12$); Ch4 is a beating pattern from $T\!=\!24$ and $T\!=\!20$ superposition.
\item \emph{Noise robustness (Ch5--Ch8).} Periodic signals from the above regimes contaminated by either white noise or red (AR(1), $\alpha\!=\!0.9$) noise.
\item \emph{No-period control (Ch9).} Pure red noise with no periodic structure, used to verify that declaring ``no period'' for a channel causes the prior to fall back to a uniform similarity matrix rather than injecting spurious bias.
\end{itemize}
A random phase shift is drawn per sample, so the ground-truth periods are invariant across samples but their alignment is not.

\subsection{Trend Dataset}
\label{app:syn-trend}

A $10$-channel dataset designed to test whether a smooth-trend prior can extrapolate a variety of curvature regimes. Time is normalised so that the history window covers $t\!\in\![0,1]$ and the forecast window covers $t\!\in\![1,2]$. The ten channels span:
\begin{itemize}
\item \emph{Linear baselines (Ch0--Ch2).} Positive-, negative-, and gentle-slope linear trends with slopes perturbed per sample.
\item \emph{Convex growth (Ch3--Ch6).} Quadratic, exponential, cubic, and noise-masked quadratic growth, testing extrapolation of a positive second derivative---including a noise-masked case where the history window visually looks linear but has non-zero curvature.
\item \emph{Concave decay (Ch7--Ch9).} Logarithmic, square-root, and saturating-inverse curves, testing extrapolation under a negative second derivative.
\end{itemize}
Each sample draws small multiplicative perturbations on the slope and curvature coefficients, and a global white-noise floor (std $0.1$) is added on top. We emphasise that these datasets are not intended as realistic benchmarks; they are controlled testbeds in which the success criterion for an RQ1 prior is whether it captures the one structural regularity the dataset isolates.

\section{Implementation Details and Hyperparameters}
\label{app:impl}

\subsection{Few-Shot Forecasting (RQ1)}

\paragraph{Synthetic datasets.} Full per-channel construction for the Lag, Periodicity, and Trend datasets is given in Appendix~\ref{app:syn-data}. In summary: Lag has $N_\text{var}\!=\!6$ channels with three causal pairs at $\tau\!=\!8$; Periodicity has $N_\text{var}\!=\!10$ channels with per-channel known periods and noise regimes; Trend has $N_\text{var}\!=\!10$ channels spanning linear baselines, convex growth, and concave decay. Each task is generated at $N\!\in\!\{150,1500,15000\}$ with a $70/10/20$ train / val / test split.

\paragraph{Training setup.} Input length $\mathrm{seq\_len}=96$; label length $\mathrm{label\_len}=48$; prediction length $\mathrm{pred\_len}=96$; $d_{\text{model}}=64$; $d_{\text{ff}}=128$; patch length $p=8$ unless otherwise stated; $h=8$ attention heads; $K=3$ MFVI iterations. Optimizer: Adam with initial learning rate $10^{-3}$ and cosine decay; batch size $32$; maximum $10$ epochs with early stopping on validation MSE. Data are instance-normalized with RevIN. All models (vanilla ST-PT, prior variants, baselines) use identical input windows and prediction horizons for a fair comparison.

\paragraph{Noise robustness sweep.} For each task, noise amplitude is introduced by adding i.i.d.\ Gaussian noise with standard deviation $\sigma_{\text{noise}}$ scaled by the per-channel signal standard deviation; $\sigma_{\text{noise}}$ takes the values given in the $x$-column of Table~\ref{tab:noise}.

\paragraph{Real-data datasets.} ETTh1, ETTh2, ETTm1, ETTm2, Weather, Exchange are used in their standard train/val/test split; short-horizon setting has $\mathrm{pred\_len}=5$, long-horizon setting has $\mathrm{pred\_len}=100$. Input length is the standard $336$ for ETT and $512$ for Weather/Exchange; other hyperparameters follow the synthetic setup.

\subsection{Condition-Programmable Factors (RQ2, PT-FG)}

\paragraph{Denoiser architecture.} Patch length $p=8$; hidden size $d_{\text{model}}=128$; topic width $d_{\text{ff}}=256$; $4$ dynamic PT blocks; $8$ heads; $16$ basis matrices for the binary topic factor and $8$ basis matrices for each of the four ternary factors (time $U,V$; channel $U,V$). The condition hidden dimension matches $d_{\text{model}}$.

\paragraph{Condition encoder.} For text-conditioned settings we consume a pre-computed LongCLIP-style text embedding of dimension $1024$. For attribute-conditioned settings each discrete attribute is embedded separately and concatenated (no multi-head cross-talk in the default best configuration). The diffusion timestep embedding is a sinusoidal encoding added to the projected condition vector.

\paragraph{Diffusion training.} $v$-prediction parameterization, quadratic $\beta$-schedule with $T=1000$ training steps (sampling uses $T_{\text{sample}}=50$ DDIM steps). Classifier-free guidance with null-condition dropout $p_{\text{null}}=0.1$ during training and guidance scale $1.5$ at inference. Optimizer: AdamW, learning rate $10^{-4}$, batch size $64$, maximum $250$ epochs with early stopping on validation loss. A spectral auxiliary loss with weight $\lambda_{\text{spec}}=0.1$ is added to the diffusion loss.

\paragraph{Compositional ablation (Table~\ref{tab:rq4}).} The four ablation variants A--D share the denoiser architecture above. Variant A inserts a single multi-head self-attention layer (with residual) among the attribute embeddings before projection; variant B uses identical embeddings but omits the self-attention layer; variant C uses only the text embedding; variant D concatenates the text embedding and the attribute embeddings and projects the result through a small MLP ($2048\!\to\!1024$). Variant D exhibits early-stopping around epoch $33$ (vs.\ $180$--$212$ for the others), indicative of training instability.

\subsection{MFVI-Based Latent-Space AR (RQ3)}

\paragraph{Architecture.} The model reuses the cornerstone's ternary and binary factors and adds: (i) a stepwise AR decoder that shares the ternary factors with the encoder but maintains its own topic modeling module; (ii) a $2$-layer MLP transition prior $p_\psi:\mathbb{R}^{d}\!\to\!\mathbb{R}^{d}$ (hidden dim $d$, GELU activation); (iii) a learnable damping coefficient $\alpha\!=\!\sigma(\text{damping\_logit})$. Hidden size $d_{\text{model}}=256$; patch length $p=8$; $e\_layers=2$ encoder MFVI iterations; $d\_layers=2$ single-step decoder iterations; $h=8$ heads. The latent consistency loss weight is $\lambda_{\text{latent}}=0.1$.

\paragraph{Training setup.} Standard long-term forecasting split on ETTh1/2, ETTm1/2, Weather, ECL. Input length $336$ for ETTh* and Weather, $512$ for ECL. Prediction lengths $\{96,192\}$. Optimizer: AdamW with learning rate $5\cdot 10^{-4}$, cosine decay, batch size $32$, maximum $50$ epochs with early stopping on validation MSE. During training, each forward pass evaluates (i) the AR student for the prediction loss and (ii) the full-sequence encoder MFVI for the teacher latents; the teacher path is under \texttt{torch.no\_grad} after the shared unary embedding.

\paragraph{Baselines.} DeepVAR, LSTM-AR, and LSTNet are trained on the same data splits with their default hyperparameters from the original implementations; AutoTimes and TCN-AR variants are excluded because their test-time metrics were not consistently available in our training logs at the time of this report.

\section{ST-PT Validation: Full Per-Horizon Results and Setting}
\label{app:cornerstone-full}

\paragraph{Setting.}
We follow the standard long-term forecasting protocol. Seven datasets are used (Weather, Electricity, Traffic, ETTh1, ETTh2, ETTm1, ETTm2) under their canonical train/val/test splits. Four forecasting horizons are evaluated: $\mathrm{pred\_len}\in\{96,192,336,720\}$. Numbers are obtained under hyperparameter search; the look-back length is searched per (dataset, model). Test MSE and Test MAE are reported; lower is better. Our method is the ST-PT cornerstone as specified in Section~\ref{sec:method}, with no RQ-specific lever activated.
\paragraph{Training and evaluation details.}
Inputs are channel-wise Z-normalized using statistics fitted on the training split and reverse-normalized before evaluation. The cornerstone hyperparameters are patch length $p\!=\!8$, hidden size $d_{\text{model}}\!=\!256$, $h\!=\!8$ attention heads per axis, $K_{\text{enc}}\!=\!2$ MFVI iterations with a learnable damping coefficient $\alpha\!=\!\sigma(\text{logit})$; RoPE is applied separately along the time and channel axes as described in Section~\ref{sec:method}. Training uses AdamW with cosine decay, batch size $32$, and MSE loss on the denormalized output, early-stopped on validation MSE with a patience of $5$ epochs. TSMixer entries are reprinted from the TSMixer paper~\cite{tsmixer}, and TimeMixer, PatchTST, TimesNet, and Crossformer entries are reprinted from the TimeMixer paper~\cite{timemixer++}, which reproduces all four under a unified protocol and supplies them as the de-facto comparison numbers for this benchmark suite.

\begin{table}[H]
\centering
\small
\setlength{\tabcolsep}{3.5pt}
\caption{Full cornerstone performace validation: Searched hyperparameter results (Test MSE / MAE) on seven datasets at four \textit{\textbf{f}orecasting \textbf{h}orizons}(fh), under hyperparameter search. The ST-PT cornerstone is compared with five strong 2023-era baselines. \textbf{TSMixer} numbers are from the TSMixer paper~\cite{tsmixer}; \textbf{TimeMixer}, \textbf{PatchTST}, \textbf{TimesNet}, and \textbf{Crossformer} numbers are from the TimeMixer paper~\cite{timemixer++}}
\label{tab:cornerstone-full}
\begin{tabular}{c c | c c | c c | c c | c c | c c | c c}
\hline
\multicolumn{2}{c|}{\multirow{2}{*}{Models}}
& \multicolumn{2}{c|}{ST-PT (Ours)}
& \multicolumn{2}{c|}{TSMixer}
& \multicolumn{2}{c|}{TimeMixer}
& \multicolumn{2}{c|}{PatchTST}
& \multicolumn{2}{c|}{TimesNet}
& \multicolumn{2}{c}{Crossformer} \\
\multicolumn{2}{c|}{} & \multicolumn{2}{c|}{} & \multicolumn{2}{c|}{2023} & \multicolumn{2}{c|}{2023} & \multicolumn{2}{c|}{2023} & \multicolumn{2}{c|}{2023} & \multicolumn{2}{c}{2023} \\
Dataset & $\mathrm{hf}$ & MSE & MAE & MSE & MAE & MSE & MAE & MSE & MAE & MSE & MAE & MSE & MAE \\
\hline\hline
\multirow{4}{*}{Weather}
& 96  & 0.159 & 0.208 & 0.145 & 0.198 & 0.147 & 0.197 & 0.149 & 0.198 & 0.172 & 0.220 & 0.232 & 0.302 \\
& 192 & 0.209 & 0.254 & 0.191 & 0.242 & 0.189 & 0.239 & 0.194 & 0.241 & 0.219 & 0.261 & 0.371 & 0.410 \\
& 336 & 0.267 & 0.295 & 0.242 & 0.280 & 0.241 & 0.280 & 0.306 & 0.282 & 0.246 & 0.337 & 0.495 & 0.515 \\
& 720 & 0.344 & 0.345 & 0.320 & 0.336 & 0.310 & 0.330 & 0.314 & 0.334 & 0.365 & 0.359 & 0.526 & 0.542 \\
\hline
\multirow{4}{*}{Electricity}
& 96  & 0.144 & 0.241 & 0.131 & 0.229 & 0.129 & 0.224 & 0.129 & 0.222 & 0.168 & 0.272 & 0.150 & 0.251 \\
& 192 & 0.160 & 0.258 & 0.151 & 0.246 & 0.140 & 0.220 & 0.147 & 0.240 & 0.184 & 0.322 & 0.161 & 0.260 \\
& 336 & 0.174 & 0.270 & 0.161 & 0.261 & 0.161 & 0.255 & 0.163 & 0.259 & 0.198 & 0.300 & 0.182 & 0.281 \\
& 720 & 0.230 & 0.321 & 0.197 & 0.293 & 0.194 & 0.287 & 0.197 & 0.290 & 0.220 & 0.320 & 0.251 & 0.339 \\
\hline
\multirow{4}{*}{Traffic}
& 96  & 0.415 & 0.277 & 0.376 & 0.264 & 0.360 & 0.249 & 0.360 & 0.249 & 0.593 & 0.321 & 0.514 & 0.267 \\
& 192 & 0.433 & 0.282 & 0.397 & 0.277 & 0.375 & 0.250 & 0.379 & 0.256 & 0.617 & 0.336 & 0.549 & 0.252 \\
& 336 & 0.452 & 0.290 & 0.413 & 0.290 & 0.385 & 0.270 & 0.392 & 0.264 & 0.629 & 0.336 & 0.530 & 0.300 \\
& 720 & 0.482 & 0.308 & 0.444 & 0.306 & 0.430 & 0.281 & 0.432 & 0.286 & 0.640 & 0.350 & 0.573 & 0.313 \\
\hline
\multirow{4}{*}{ETTh1}
& 96  & 0.376 & 0.400 & 0.361 & 0.392 & 0.361 & 0.390 & 0.370 & 0.400 & 0.384 & 0.402 & 0.418 & 0.438 \\
& 192 & 0.428 & 0.426 & 0.404 & 0.418 & 0.409 & 0.414 & 0.413 & 0.429 & 0.436 & 0.429 & 0.539 & 0.517 \\
& 336 & 0.480 & 0.450 & 0.420 & 0.431 & 0.430 & 0.429 & 0.422 & 0.440 & 0.638 & 0.469 & 0.709 & 0.638 \\
& 720 & 0.464 & 0.460 & 0.463 & 0.472 & 0.445 & 0.460 & 0.447 & 0.468 & 0.521 & 0.500 & 0.733 & 0.636 \\
\hline
\multirow{4}{*}{ETTh2}
& 96  & 0.278 & 0.335 & 0.274 & 0.341 & 0.271 & 0.330 & 0.274 & 0.337 & 0.340 & 0.374 & 0.425 & 0.463 \\
& 192 & 0.370 & 0.396 & 0.339 & 0.385 & 0.317 & 0.402 & 0.314 & 0.382 & 0.231 & 0.322 & 0.473 & 0.500 \\
& 336 & 0.408 & 0.424 & 0.361 & 0.406 & 0.332 & 0.396 & 0.329 & 0.384 & 0.452 & 0.452 & 0.581 & 0.562 \\
& 720 & 0.417 & 0.440 & 0.445 & 0.470 & 0.342 & 0.408 & 0.379 & 0.422 & 0.462 & 0.468 & 0.775 & 0.665 \\
\hline
\multirow{4}{*}{ETTm1}
& 96  & 0.315 & 0.358 & 0.285 & 0.339 & 0.291 & 0.340 & 0.293 & 0.346 & 0.338 & 0.375 & 0.361 & 0.403 \\
& 192 & 0.381 & 0.395 & 0.327 & 0.365 & 0.327 & 0.365 & 0.333 & 0.370 & 0.374 & 0.387 & 0.387 & 0.422 \\
& 336 & 0.406 & 0.412 & 0.356 & 0.382 & 0.360 & 0.381 & 0.369 & 0.392 & 0.410 & 0.411 & 0.605 & 0.572 \\
& 720 & 0.468 & 0.448 & 0.419 & 0.414 & 0.415 & 0.417 & 0.416 & 0.420 & 0.478 & 0.450 & 0.703 & 0.645 \\
\hline
\multirow{4}{*}{ETTm2}
& 96  & 0.173 & 0.261 & 0.163 & 0.252 & 0.164 & 0.254 & 0.166 & 0.256 & 0.187 & 0.267 & 0.275 & 0.358 \\
& 192 & 0.245 & 0.308 & 0.216 & 0.290 & 0.223 & 0.295 & 0.223 & 0.296 & 0.249 & 0.309 & 0.345 & 0.400 \\
& 336 & 0.306 & 0.348 & 0.268 & 0.324 & 0.279 & 0.330 & 0.274 & 0.329 & 0.321 & 0.351 & 0.657 & 0.528 \\
& 720 & 0.409 & 0.410 & 0.420 & 0.422 & 0.359 & 0.383 & 0.362 & 0.385 & 0.408 & 0.403 & 1.208 & 0.753 \\
\hline
\end{tabular}
\end{table}

\section{ConTSG-Bench Per-Dataset Raw Results}
\label{app:contsg-raw}

This appendix provides the raw (per-dataset) numerical results that back the rank-based summary in Table~\ref{tab:contsg-rank}. All metrics are lower-is-better. Baseline entries are mean$\pm$sample-std across three random seeds; PT-FG entries are mean$\pm$sample-std across seeds $\{42,142,242\}$. A dash (---) indicates that the corresponding metric could not be recovered from the log. All ten tables are presented together for ease of cross-referencing. Apart from PT-FG (ours), all baseline numbers in the tables below are reprinted from the ConTSG-Bench paper~\cite{contsgbench2026}.

\begin{table}[H]
\centering
\footnotesize
\caption{Synth-U.}
\label{tab:app-synthu}
\begin{tabular}{lcccccc}
\toprule
Model & DTW & CRPS & ACD & SD & KD & MDD \\
\midrule
Bridge        & 5.884$\pm$0.404 & 0.517$\pm$0.015 & 0.031$\pm$0.012 & 0.097$\pm$0.078 & 0.642$\pm$0.056 & 0.028$\pm$0.000 \\
DiffuSETS     & 5.684$\pm$0.800 & 0.439$\pm$0.016 & 0.055$\pm$0.011 & 0.140$\pm$0.138 & 0.833$\pm$0.865 & 0.020$\pm$0.005 \\
T2S           & 27.264$\pm$17.270 & 1.926$\pm$1.364 & 0.049$\pm$0.010 & 0.496$\pm$0.411 & 0.783$\pm$0.430 & 0.036$\pm$0.017 \\
TEdit         & 6.928$\pm$1.026 & 0.631$\pm$0.140 & 0.066$\pm$0.001 & 0.163$\pm$0.107 & 0.288$\pm$0.115 & 0.022$\pm$0.006 \\
Text2Motion   & 5.048$\pm$0.088 & 0.435$\pm$0.016 & 0.082$\pm$0.007 & 0.072$\pm$0.041 & 0.205$\pm$0.067 & 0.015$\pm$0.003 \\
TimeVQVAE     & 4.773$\pm$0.004 & 0.461$\pm$0.002 & 0.080$\pm$0.002 & 0.046$\pm$0.024 & 0.776$\pm$0.017 & 0.025$\pm$0.001 \\
TimeWeaver    & 5.982$\pm$0.755 & 0.548$\pm$0.104 & 0.074$\pm$0.010 & 0.047$\pm$0.018 & 0.521$\pm$0.177 & 0.021$\pm$0.006 \\
TTSCGAN       & 8.690$\pm$0.068 & 0.636$\pm$0.017 & 0.285$\pm$0.002 & 0.104$\pm$0.058 & 0.135$\pm$0.073 & 0.023$\pm$0.005 \\
VerbalTS      & 4.885$\pm$0.159 & 0.448$\pm$0.017 & 0.060$\pm$0.004 & 0.032$\pm$0.025 & 0.374$\pm$0.060 & 0.016$\pm$0.002 \\
WaveStitch    & 13.837$\pm$7.071 & 1.181$\pm$0.619 & 0.049$\pm$0.026 & 0.124$\pm$0.110 & 0.340$\pm$0.163 & 0.035$\pm$0.014 \\
PT-FG (ours)  & 4.842$\pm$0.031 & 0.433$\pm$0.002 & 0.073$\pm$0.000 & 0.050$\pm$0.005 & 0.342$\pm$0.004 & 0.013$\pm$0.000 \\
\bottomrule
\end{tabular}
\end{table}
\vspace{-20pt}
\begin{table}[H]
\centering
\footnotesize
\caption{Synth-M.}
\label{tab:app-synthm}
\begin{tabular}{lcccccc}
\toprule
Model & DTW & CRPS & ACD & SD & KD & MDD \\
\midrule
Bridge        & 12.558$\pm$1.073 & 0.596$\pm$0.025 & 0.058$\pm$0.024 & 0.138$\pm$0.170 & 0.422$\pm$0.247 & 0.029$\pm$0.000 \\
DiffuSETS     & 10.234$\pm$0.851 & 0.458$\pm$0.022 & 0.041$\pm$0.019 & 0.085$\pm$0.071 & 0.160$\pm$0.141 & 0.021$\pm$0.003 \\
T2S           & 38.741$\pm$11.492 & 1.678$\pm$0.682 & 0.075$\pm$0.036 & 0.397$\pm$0.117 & 1.495$\pm$0.172 & 0.031$\pm$0.005 \\
TEdit         & 9.978$\pm$0.629 & 0.466$\pm$0.023 & 0.056$\pm$0.006 & 0.031$\pm$0.007 & 0.255$\pm$0.109 & 0.016$\pm$0.002 \\
Text2Motion   & 9.777$\pm$0.177 & 0.441$\pm$0.007 & 0.076$\pm$0.013 & 0.057$\pm$0.046 & 0.248$\pm$0.047 & 0.014$\pm$0.001 \\
TimeVQVAE     & 9.243$\pm$0.188 & 0.488$\pm$0.013 & 0.073$\pm$0.003 & 0.028$\pm$0.015 & 0.631$\pm$0.113 & 0.021$\pm$0.001 \\
TimeWeaver    & 9.730$\pm$0.334 & 0.466$\pm$0.031 & 0.058$\pm$0.007 & 0.051$\pm$0.021 & 0.176$\pm$0.154 & 0.016$\pm$0.001 \\
TTSCGAN       & 13.451$\pm$0.169 & 0.658$\pm$0.012 & 0.265$\pm$0.000 & 0.099$\pm$0.046 & 0.645$\pm$0.032 & 0.040$\pm$0.001 \\
VerbalTS      & 10.118$\pm$0.652 & 0.474$\pm$0.053 & 0.051$\pm$0.002 & 0.041$\pm$0.032 & 0.193$\pm$0.060 & 0.015$\pm$0.000 \\
WaveStitch    & 16.094$\pm$2.092 & 0.784$\pm$0.103 & 0.054$\pm$0.023 & 0.215$\pm$0.134 & 0.303$\pm$0.233 & 0.014$\pm$0.004 \\
PT-FG (ours)  & 9.321$\pm$0.019 & 0.431$\pm$0.001 & 0.051$\pm$0.000 & 0.062$\pm$0.002 & 0.358$\pm$0.012 & 0.011$\pm$0.000 \\
\bottomrule
\end{tabular}
\end{table}
\vspace{-20pt}
\begin{table}[H]
\centering
\footnotesize
\caption{AirQuality Beijing.}
\label{tab:app-air}
\begin{tabular}{lcccccc}
\toprule
Model & DTW & CRPS & ACD & SD & KD & MDD \\
\midrule
Bridge        & 7.078$\pm$0.209 & 0.327$\pm$0.010 & 0.034$\pm$0.007 & 0.209$\pm$0.056 & 0.737$\pm$0.069 & 0.029$\pm$0.001 \\
DiffuSETS     & 6.371$\pm$0.111 & 0.309$\pm$0.005 & 0.047$\pm$0.005 & 0.119$\pm$0.017 & 0.360$\pm$0.118 & 0.025$\pm$0.002 \\
T2S           & 29.786$\pm$18.323 & 1.644$\pm$0.808 & 0.108$\pm$0.026 & 0.348$\pm$0.144 & 1.535$\pm$0.262 & 0.038$\pm$0.004 \\
TEdit         & 6.757$\pm$0.106 & 0.305$\pm$0.003 & 0.037$\pm$0.008 & 0.098$\pm$0.017 & 0.514$\pm$0.082 & 0.021$\pm$0.001 \\
Text2Motion   & 5.947$\pm$0.022 & 0.314$\pm$0.001 & 0.052$\pm$0.002 & 0.144$\pm$0.066 & 0.500$\pm$0.019 & 0.025$\pm$0.000 \\
TimeVQVAE     & 6.481 & 0.318 & --- & --- & --- & 0.047$\pm$0.020 \\
TimeWeaver    & 6.971$\pm$0.469 & 0.310$\pm$0.015 & 0.040$\pm$0.009 & 0.366$\pm$0.372 & 1.116$\pm$0.938 & 0.022$\pm$0.001 \\
TTSCGAN       & 7.779$\pm$0.100 & 0.427$\pm$0.012 & 0.145$\pm$0.001 & 0.355$\pm$0.026 & 1.139$\pm$0.085 & 0.041$\pm$0.002 \\
VerbalTS      & 6.731$\pm$0.111 & 0.308$\pm$0.008 & 0.043$\pm$0.007 & 0.166$\pm$0.062 & 0.646$\pm$0.076 & 0.024$\pm$0.002 \\
WaveStitch    & 8.196$\pm$0.420 & 0.401$\pm$0.044 & 0.050$\pm$0.006 & 0.448$\pm$0.074 & 3.699$\pm$2.380 & 0.026$\pm$0.002 \\
PT-FG (ours)  & 6.259$\pm$0.011 & 0.287$\pm$0.000 & 0.039$\pm$0.000 & 0.242$\pm$0.005 & 0.581$\pm$0.010 & 0.019$\pm$0.000 \\
\bottomrule
\end{tabular}
\end{table}
\vspace{-20pt}
\begin{table}[H]
\centering
\footnotesize
\caption{ETTm1.}
\label{tab:app-ettm1}
\begin{tabular}{lcccccc}
\toprule
Model & DTW & CRPS & ACD & SD & KD & MDD \\
\midrule
Bridge        & 8.080$\pm$0.171 & 0.550$\pm$0.009 & 0.107$\pm$0.021 & 0.493$\pm$0.032 & 3.352$\pm$0.066 & 0.027$\pm$0.002 \\
DiffuSETS     & 8.200$\pm$0.596 & 0.592$\pm$0.093 & 0.142$\pm$0.086 & 8.533$\pm$11.365 & 258.72$\pm$428.35 & 0.042$\pm$0.013 \\
T2S           & 69.319$\pm$39.256 & 5.395$\pm$3.403 & 0.115$\pm$0.132 & 1.127$\pm$1.530 & 3.133$\pm$1.877 & 0.054$\pm$0.006 \\
TEdit         & 7.875$\pm$0.057 & 0.539$\pm$0.020 & 0.169$\pm$0.050 & 0.749$\pm$0.185 & 3.879$\pm$0.398 & 0.023$\pm$0.006 \\
Text2Motion   & 8.560$\pm$0.596 & 0.530$\pm$0.035 & 0.111$\pm$0.042 & 0.865$\pm$1.188 & 8.836$\pm$13.631 & 0.024$\pm$0.005 \\
TimeVQVAE     & 7.540$\pm$0.060 & 0.537$\pm$0.010 & 0.053$\pm$0.018 & 1.952$\pm$0.199 & 9.304$\pm$5.151 & 0.036$\pm$0.003 \\
TimeWeaver    & 10.463$\pm$1.117 & 0.723$\pm$0.108 & 0.084$\pm$0.042 & 1.031$\pm$0.202 & 4.196$\pm$0.916 & 0.028$\pm$0.010 \\
TTSCGAN       & 11.116$\pm$3.434 & 0.863$\pm$0.268 & 0.303$\pm$0.000 & 1.486$\pm$0.130 & 4.867$\pm$0.074 & 0.040$\pm$0.013 \\
VerbalTS      & 9.728$\pm$2.634 & 0.668$\pm$0.168 & 0.158$\pm$0.085 & 1.784$\pm$1.009 & 4.599$\pm$2.960 & 0.031$\pm$0.016 \\
WaveStitch    & 9.853$\pm$1.112 & 0.662$\pm$0.100 & 0.052$\pm$0.026 & 0.810$\pm$0.684 & 2.110$\pm$1.826 & 0.035$\pm$0.004 \\
PT-FG (ours)  & 6.917$\pm$0.048 & 0.459$\pm$0.004 & 0.043$\pm$0.003 & 0.625$\pm$0.160 & 0.607$\pm$0.224 & 0.015$\pm$0.000 \\
\bottomrule
\end{tabular}
\end{table}
\vspace{-20pt}
\begin{table}[H]
\centering
\footnotesize
\caption{Istanbul Traffic.}
\label{tab:app-istanbul}
\begin{tabular}{lcccccc}
\toprule
Model & DTW & CRPS & ACD & SD & KD & MDD \\
\midrule
Bridge        & 8.298$\pm$1.982 & 0.525$\pm$0.070 & 0.029$\pm$0.002 & 0.481$\pm$0.256 & 0.150$\pm$0.194 & 0.036$\pm$0.005 \\
DiffuSETS     & 8.571$\pm$3.957 & 0.528$\pm$0.255 & 0.147$\pm$0.088 & 2.235$\pm$1.591 & 12.996$\pm$11.174 & 0.041$\pm$0.019 \\
T2S           & 21.861$\pm$8.202 & 1.126$\pm$0.417 & 0.245$\pm$0.033 & 0.319$\pm$0.346 & 1.518$\pm$1.070 & 0.028$\pm$0.007 \\
TEdit         & 8.722$\pm$0.351 & 0.496$\pm$0.003 & 0.031$\pm$0.001 & 0.289$\pm$0.183 & 0.299$\pm$0.228 & 0.018$\pm$0.002 \\
Text2Motion   & 3.748$\pm$0.197 & 0.260$\pm$0.005 & 0.057$\pm$0.010 & 0.145$\pm$0.080 & 0.078$\pm$0.075 & 0.010$\pm$0.002 \\
TimeVQVAE     & 9.824$\pm$0.869 & 0.693$\pm$0.036 & 0.202$\pm$0.200 & 0.322$\pm$0.126 & 2.062$\pm$2.235 & 0.040$\pm$0.005 \\
TimeWeaver    & 8.927$\pm$0.326 & 0.498$\pm$0.011 & 0.027$\pm$0.004 & 0.206$\pm$0.146 & 0.515$\pm$0.159 & 0.020$\pm$0.001 \\
TTSCGAN       & 12.090$\pm$0.943 & 0.649$\pm$0.063 & 0.340$\pm$0.002 & 0.110$\pm$0.173 & 0.316$\pm$0.382 & 0.021$\pm$0.006 \\
VerbalTS      & 3.936$\pm$0.724 & 0.250$\pm$0.031 & 0.024$\pm$0.006 & 0.132$\pm$0.175 & 0.092$\pm$0.155 & 0.011$\pm$0.001 \\
WaveStitch    & 10.775$\pm$2.371 & 0.640$\pm$0.229 & 0.022$\pm$0.014 & 0.496$\pm$0.509 & 0.582$\pm$0.823 & 0.018$\pm$0.009 \\
PT-FG (ours)  & 3.458$\pm$0.013 & 0.234$\pm$0.001 & 0.039$\pm$0.001 & 0.065$\pm$0.006 & 0.105$\pm$0.010 & 0.010$\pm$0.000 \\
\bottomrule
\end{tabular}
\end{table}
\vspace{-20pt}
\begin{table}[H]
\centering
\footnotesize
\caption{TelecomTS.}
\label{tab:app-telecom}
\begin{tabular}{lcccccc}
\toprule
Model & DTW & CRPS & ACD & SD & KD & MDD \\
\midrule
Bridge        & 13.805$\pm$0.361 & 0.537$\pm$0.005 & 0.049$\pm$0.011 & 0.949$\pm$0.400 & 9.113$\pm$2.349 & 0.033$\pm$0.009 \\
DiffuSETS     & 9.908$\pm$1.026 & 0.367$\pm$0.041 & 0.051$\pm$0.023 & 0.726$\pm$0.440 & 5.620$\pm$2.334 & 0.024$\pm$0.006 \\
T2S           & 44.384$\pm$24.553 & 1.831$\pm$0.929 & 0.104$\pm$0.046 & 0.511$\pm$0.363 & 7.417$\pm$1.474 & 0.035$\pm$0.010 \\
TEdit         & 14.355$\pm$0.785 & 0.524$\pm$0.025 & 0.066$\pm$0.020 & 0.519$\pm$0.270 & 8.374$\pm$1.404 & 0.024$\pm$0.003 \\
Text2Motion   & 8.728$\pm$0.233 & 0.295$\pm$0.011 & 0.079$\pm$0.019 & 0.990$\pm$0.519 & 6.903$\pm$7.445 & 0.013$\pm$0.001 \\
TimeVQVAE     & 14.029$\pm$0.131 & 0.615$\pm$0.008 & 0.094$\pm$0.005 & 0.622$\pm$0.541 & 9.242$\pm$5.111 & 0.036$\pm$0.004 \\
TimeWeaver    & 14.714$\pm$0.785 & 0.557$\pm$0.039 & 0.040$\pm$0.002 & 0.513$\pm$0.436 & 7.886$\pm$1.046 & 0.027$\pm$0.003 \\
TTSCGAN       & 14.068$\pm$0.484 & 0.599$\pm$0.038 & 0.097$\pm$0.002 & 0.677$\pm$0.112 & 7.160$\pm$0.068 & 0.034$\pm$0.005 \\
VerbalTS      & 11.878$\pm$5.174 & 0.426$\pm$0.190 & 0.068$\pm$0.012 & 0.746$\pm$0.457 & 3.051$\pm$2.188 & 0.024$\pm$0.012 \\
WaveStitch    & 15.179$\pm$0.892 & 0.573$\pm$0.044 & 0.036$\pm$0.013 & 9.114$\pm$7.850 & 520.83$\pm$673.31 & 0.025$\pm$0.003 \\
PT-FG (ours)  & 8.698$\pm$0.011 & 0.288$\pm$0.000 & 0.088$\pm$0.001 & 1.656$\pm$0.013 & 6.501$\pm$0.178 & 0.011$\pm$0.000 \\
\bottomrule
\end{tabular}
\end{table}
\vspace{-20pt}
\begin{table}[H]
\centering
\footnotesize
\caption{Weather Conceptual.}
\label{tab:app-weather-c}
\begin{tabular}{lcccccc}
\toprule
Model & DTW & CRPS & ACD & SD & KD & MDD \\
\midrule
Bridge        & 15.717$\pm$0.663 & 0.405$\pm$0.027 & 0.034$\pm$0.003 & 19.567$\pm$0.089 & 2320.63$\pm$0.56 & 0.021$\pm$0.000 \\
DiffuSETS     & 67.225$\pm$64.594 & 1.136$\pm$0.715 & 0.138$\pm$0.060 & 19.121$\pm$0.273 & 2317.85$\pm$1.54 & 0.040$\pm$0.000 \\
T2S           & 128.763$\pm$112.845 & 4.150$\pm$3.606 & 0.093$\pm$0.010 & 18.813$\pm$0.222 & 2315.78$\pm$3.18 & 0.030$\pm$0.012 \\
TEdit         & 12.136$\pm$0.175 & 0.254$\pm$0.006 & 0.034$\pm$0.008 & 23.440$\pm$3.557 & 2362.11$\pm$232.11 & 0.009$\pm$0.001 \\
Text2Motion   & 10.292$\pm$0.109 & 0.251$\pm$0.002 & 0.050$\pm$0.007 & 18.120$\pm$0.091 & 2298.91$\pm$3.65 & 0.008$\pm$0.000 \\
TimeVQVAE     & 15.655$\pm$0.088 & 0.451$\pm$0.009 & 0.071$\pm$0.016 & 19.034$\pm$0.117 & 2319.52$\pm$4.11 & 0.026$\pm$0.001 \\
TimeWeaver    & 12.558$\pm$0.807 & 0.263$\pm$0.019 & 0.042$\pm$0.011 & 30.757$\pm$5.029 & 2618.78$\pm$427.46 & 0.009$\pm$0.001 \\
TTSCGAN       & 18.131$\pm$0.205 & 0.507$\pm$0.002 & 0.167$\pm$0.001 & 19.557$\pm$0.047 & 2319.85$\pm$0.02 & 0.028$\pm$0.001 \\
VerbalTS      & 10.996$\pm$0.127 & 0.219$\pm$0.003 & 0.031$\pm$0.006 & 16.877$\pm$0.325 & 2119.48$\pm$115.59 & 0.007$\pm$0.001 \\
WaveStitch    & 15.176$\pm$3.057 & 0.317$\pm$0.059 & 0.036$\pm$0.008 & 19.340$\pm$3.042 & 2381.97$\pm$223.82 & 0.010$\pm$0.002 \\
PT-FG (ours)  & 10.289$\pm$0.004 & 0.210$\pm$0.001 & 0.071$\pm$0.000 & 17.566$\pm$0.052 & 2279.02$\pm$1.64 & 0.006$\pm$0.000 \\
\bottomrule
\end{tabular}
\end{table}
\vspace{-20pt}
\begin{table}[H]
\centering
\footnotesize
\caption{Weather Morphological.}
\label{tab:app-weather-m}
\begin{tabular}{lcccccc}
\toprule
Model & DTW & CRPS & ACD & SD & KD & MDD \\
\midrule
Bridge        & 17.786$\pm$0.587 & 0.458$\pm$0.015 & 0.043$\pm$0.006 & 20.214$\pm$1.094 & 2332.32$\pm$23.79 & 0.025$\pm$0.002 \\
DiffuSETS     & 26.874$\pm$11.800 & 0.804$\pm$0.422 & 0.114$\pm$0.019 & 19.397$\pm$0.096 & 2316.96$\pm$0.88 & 0.035$\pm$0.007 \\
T2S           & 256.791$\pm$87.216 & 7.830$\pm$2.999 & 0.105$\pm$0.006 & 18.938$\pm$0.118 & 2307.18$\pm$15.18 & 0.038$\pm$0.006 \\
TEdit         & 19.165$\pm$0.251 & 0.450$\pm$0.035 & 0.068$\pm$0.015 & 22.165$\pm$2.866 & 2419.35$\pm$169.24 & 0.022$\pm$0.003 \\
Text2Motion   & 13.600$\pm$0.079 & 0.308$\pm$0.006 & 0.054$\pm$0.001 & 16.946$\pm$0.889 & 2202.26$\pm$81.51 & 0.012$\pm$0.001 \\
TimeVQVAE     & 18.060$\pm$0.268 & 0.497$\pm$0.005 & 0.074$\pm$0.011 & 19.500$\pm$0.035 & 2320.32$\pm$5.65 & 0.028$\pm$0.001 \\
TimeWeaver    & 18.116$\pm$0.483 & 0.427$\pm$0.007 & 0.058$\pm$0.005 & 30.500$\pm$9.880 & 3074.84$\pm$1242.87 & 0.021$\pm$0.003 \\
TTSCGAN       & 17.945$\pm$0.126 & 0.504$\pm$0.002 & 0.167$\pm$0.000 & 19.579$\pm$0.034 & 2317.64$\pm$0.01 & 0.029$\pm$0.002 \\
VerbalTS      & 15.182$\pm$0.851 & 0.332$\pm$0.020 & 0.038$\pm$0.010 & 19.947$\pm$5.128 & 2189.22$\pm$79.79 & 0.013$\pm$0.002 \\
WaveStitch    & 26.229$\pm$6.257 & 0.602$\pm$0.131 & 0.085$\pm$0.028 & 23.566$\pm$3.567 & 2753.79$\pm$388.61 & 0.023$\pm$0.001 \\
PT-FG (ours)  & 14.330$\pm$0.026 & 0.311$\pm$0.001 & 0.060$\pm$0.001 & 16.750$\pm$0.842 & 2240.31$\pm$29.85 & 0.012$\pm$0.000 \\
\bottomrule
\end{tabular}
\end{table}
\vspace{-20pt}
\begin{table}[H]
\centering
\footnotesize
\caption{PTB-XL Conceptual.}
\label{tab:app-ptb-c}
\begin{tabular}{lcccccc}
\toprule
Model & DTW & CRPS & ACD & SD & KD & MDD \\
\midrule
Bridge        & 97.351$\pm$0.157 & 0.426$\pm$0.003 & 0.062$\pm$0.011 & 2.810$\pm$0.021 & 184.41$\pm$0.23 & 0.015$\pm$0.000 \\
DiffuSETS     & 311.484$\pm$369.442 & 0.805$\pm$0.651 & 0.105$\pm$0.016 & 5.813$\pm$2.084 & 212.32$\pm$45.91 & 0.019$\pm$0.011 \\
T2S           & 176.159$\pm$65.460 & 0.921$\pm$0.410 & 0.211$\pm$0.047 & 3.212$\pm$0.204 & 177.64$\pm$4.77 & 0.030$\pm$0.011 \\
TEdit         & 101.870$\pm$4.600 & 0.453$\pm$0.023 & 0.086$\pm$0.017 & 2.786$\pm$0.025 & 182.95$\pm$0.45 & 0.012$\pm$0.003 \\
Text2Motion   & 100.853$\pm$0.046 & 0.514$\pm$0.008 & 0.323$\pm$0.030 & 3.652$\pm$0.325 & 176.50$\pm$6.79 & 0.018$\pm$0.000 \\
TimeVQVAE     & 98.303$\pm$0.047 & 0.438$\pm$0.001 & 0.133$\pm$0.007 & 3.069$\pm$0.076 & 178.43$\pm$5.58 & 0.017$\pm$0.000 \\
TimeWeaver    & 136.767$\pm$6.184 & 0.729$\pm$0.028 & 0.096$\pm$0.013 & 2.722$\pm$0.089 & 182.53$\pm$1.08 & 0.017$\pm$0.002 \\
TTSCGAN       & 99.063$\pm$0.723 & 0.456$\pm$0.012 & 0.119$\pm$0.005 & 2.894$\pm$0.017 & 185.13$\pm$0.09 & 0.010$\pm$0.001 \\
VerbalTS      & 97.891$\pm$3.039 & 0.431$\pm$0.011 & 0.088$\pm$0.004 & 2.772$\pm$0.039 & 182.90$\pm$0.43 & 0.015$\pm$0.002 \\
WaveStitch    & 448.352$\pm$70.124 & 2.360$\pm$0.270 & 0.246$\pm$0.023 & 2.917$\pm$0.134 & 183.03$\pm$1.03 & 0.045$\pm$0.003 \\
PT-FG (ours)  & 98.375$\pm$0.005 & 0.418$\pm$0.000 & 0.083$\pm$0.000 & 2.827$\pm$0.019 & 178.54$\pm$0.46 & 0.015$\pm$0.000 \\
\bottomrule
\end{tabular}
\end{table}
\vspace{-20pt}
\begin{table}[H]
\centering
\footnotesize
\caption{PTB-XL Morphological.}
\label{tab:app-ptb-m}
\begin{tabular}{lcccccc}
\toprule
Model & DTW & CRPS & ACD & SD & KD & MDD \\
\midrule
Bridge        & 97.589$\pm$0.276 & 0.432$\pm$0.003 & 0.056$\pm$0.007 & 2.816$\pm$0.026 & 184.28$\pm$0.09 & 0.016$\pm$0.000 \\
DiffuSETS     & 167.254$\pm$113.459 & 0.721$\pm$0.439 & 0.107$\pm$0.021 & 2.643$\pm$0.443 & 157.96$\pm$33.02 & 0.013$\pm$0.005 \\
T2S           & 152.616$\pm$86.597 & 0.890$\pm$0.531 & 0.194$\pm$0.034 & 3.599$\pm$0.988 & 176.71$\pm$6.98 & 0.030$\pm$0.015 \\
TEdit         & 100.344$\pm$3.766 & 0.459$\pm$0.044 & 0.090$\pm$0.007 & 2.819$\pm$0.044 & 182.79$\pm$0.41 & 0.011$\pm$0.001 \\
Text2Motion   & 100.645$\pm$0.033 & 0.519$\pm$0.006 & 0.250$\pm$0.083 & 4.743$\pm$0.114 & 165.22$\pm$3.72 & 0.019$\pm$0.000 \\
TimeVQVAE     & 98.278$\pm$0.106 & 0.440$\pm$0.002 & 0.127$\pm$0.013 & 3.856$\pm$0.708 & 204.66$\pm$31.98 & 0.017$\pm$0.000 \\
TimeWeaver    & 200.199$\pm$82.203 & 0.933$\pm$0.324 & 0.158$\pm$0.074 & 2.904$\pm$0.180 & 182.14$\pm$1.19 & 0.028$\pm$0.014 \\
TTSCGAN       & 99.629$\pm$1.939 & 0.469$\pm$0.012 & 0.114$\pm$0.011 & 2.895$\pm$0.090 & 185.07$\pm$0.16 & 0.010$\pm$0.001 \\
VerbalTS      & 98.542$\pm$3.557 & 0.430$\pm$0.023 & 0.081$\pm$0.001 & 2.748$\pm$0.073 & 182.23$\pm$0.84 & 0.015$\pm$0.002 \\
WaveStitch    & 344.872$\pm$168.206 & 1.912$\pm$0.994 & 0.175$\pm$0.057 & 2.912$\pm$0.174 & 182.51$\pm$0.51 & 0.042$\pm$0.009 \\
PT-FG (ours)  & 98.198$\pm$0.007 & 0.416$\pm$0.000 & 0.075$\pm$0.000 & 2.786$\pm$0.031 & 178.52$\pm$0.56 & 0.014$\pm$0.000 \\
\bottomrule
\end{tabular}
\end{table}

\section{Pseudocode for the MFVI-Based Latent-Space AR Model}
\label{app:latent-ar-pseudo}

This appendix gives the full computation flow of the RQ3 model. We distinguish training (teacher + student) and inference (student only). Both share the same MFVI operator and the same factor matrices; training differs from inference only in that the teacher path runs an extra full-sequence encoder MFVI on history-plus-ground-truth-future to provide oracle latents for distillation.

\paragraph{Notation.}
$\mathbf{X}_{\mathrm{hist}}\!\in\!\mathbb{R}^{B\times N\times P\times p}$ denotes the history tensor ($B$ batch, $N$ channels, $P$ history patches, $p$ patch length); $\mathbf{Y}\!\in\!\mathbb{R}^{B\times N\times P_f\times p}$ is the ground-truth future where $P_f\!=\!S_f/p$ is the number of future patches. $\phi_u$ is the shared unary embedding, $\{U^{\text{time}},V^{\text{time}},U^{\text{chan}},V^{\text{chan}}\}$ are the ternary factors, $\phi_b$ the binary topic factor, $p_\psi$ the transition-prior MLP, $\alpha\!\in\!(0,1)$ the learnable damping, and $\mathrm{MLP}_{\mathrm{pred}}$ the patch-prediction head. $\mathrm{EncoderMFVI}(\cdot;K)$ runs $K$ rounds of the ST-PT update~\eqref{eq:H}--\eqref{eq:Z} on its input, \emph{with a causal mask on the time axis}: position $t$ attends only to $s\!\le\!t$, realised by an additive $-\infty$ mask on the upper triangle of the time-axis message-$F$ logits. This differs from the cornerstone encoder of Section~\ref{sec:method} (which is non-causal along time); the RQ3 setting specifically uses a causal variant of the same operator, both for the teacher (on $[\mathbf{X}_{\mathrm{hist}}\,\|\,\mathbf{Y}]$) and for the student (on $\mathbf{X}_{\mathrm{hist}}$). $\mathrm{SingleStepMFVI}(\cdot)$ is one round of the same causal update restricted to a single new time slice. We write $\mathbf{Z}[{:,:,t,:}]$ for the slice at patch index $t$.

\paragraph{Reading notes.}
Four places in the algorithms implement the key design choices of RQ3. First, line~11 of Algorithm~\ref{alg:latent-ar-train} (and line~7 of Algorithm~\ref{alg:latent-ar-infer}) shows the \emph{dual-pathway unary}: the next-slice latent receives both a re-encoded prediction (committing to the observation space) and a latent-space shortcut from the transition prior. Second, the call to $\mathrm{SingleStepMFVI}$ is one round of the same causal ST-PT update applied to the new slice only, so the latent transition is a Bayesian posterior update rather than a stand-alone MLP. Third, the teacher in Algorithm~\ref{alg:latent-ar-train} differs from the student \emph{only} in its input---full sequence vs.\ history only---and uses the same MFVI operator, so the distillation loss compares posteriors that live in the same belief space. Fourth, because the encoder MFVI is causal along the time axis (see notation), the teacher's history-position latents are identical to the student's: $\mathbf{Z}^{\mathrm{teacher}}[:,:,{:}P,:]\!\equiv\!\mathbf{Z}^{\mathrm{stu}}[:,:,{:}P,:]$; the distillation loss therefore acts exclusively on the $P_f$ future positions, where the two paths differ only in their future-patch evidence (ground-truth for the teacher, self-rolled for the student).

\begin{algorithm}[H]
\caption{Inference (rollout) of the MFVI-based latent-space AR model.}
\label{alg:latent-ar-infer}
\begin{algorithmic}[1]
\Require history $\mathbf{X}_{\mathrm{hist}}$, trained parameters $\Theta$
\Ensure prediction $\hat{\mathbf{Y}}$
\State $\mathbf{Z}^{\mathrm{stu}}\gets \mathrm{EncoderMFVI}\!\big(\phi_u(\mathbf{X}_{\mathrm{hist}});\,K_{\mathrm{enc}}\big)$ \Comment{history-only cache}
\State $\hat{\mathbf{Y}}\gets []$
\For{$k = 1,\dots, P_f$}
  \State $z_{\mathrm{prev}}\gets \mathbf{Z}^{\mathrm{stu}}[:,:,P+k-1,:]$
  \State $\hat{\mathbf{x}}_{k}\gets \mathrm{MLP}_{\mathrm{pred}}(z_{\mathrm{prev}})$;\quad append to $\hat{\mathbf{Y}}$
  \State $u_{\mathrm{patch}}\gets \phi_u(\hat{\mathbf{x}}_{k})$;\quad $u_{\mathrm{trans}}\gets p_\psi(z_{\mathrm{prev}})$
  \State $z_{\mathrm{new}}\gets \mathrm{SingleStepMFVI}\!\big(u_{\mathrm{patch}}+u_{\mathrm{trans}},\;\mathbf{Z}^{\mathrm{stu}},\;\phi_b;\,K_{\mathrm{dec}}\big)$
  \State append $\alpha\cdot z_{\mathrm{prev}} + (1-\alpha)\cdot z_{\mathrm{new}}$ to $\mathbf{Z}^{\mathrm{stu}}$
\EndFor
\State \Return $\hat{\mathbf{Y}}$
\end{algorithmic}
\end{algorithm}

\begin{algorithm}[H]
\caption{Training step of the MFVI-based latent-space AR model.}
\label{alg:latent-ar-train}
\begin{algorithmic}[1]
\Require history $\mathbf{X}_{\mathrm{hist}}$, future $\mathbf{Y}$; parameters $\Theta=\{\phi_u,U^{\cdot},V^{\cdot},\phi_b,p_\psi,\alpha,\mathrm{MLP}_{\mathrm{pred}}\}$
\Ensure loss $\mathcal{L}$
\State \textit{\# Teacher path (no gradient after unary embedding)}
\State $\mathbf{Z}^{\mathrm{full}}\gets \phi_u\!\big([\,\mathbf{X}_{\mathrm{hist}}\,\|\,\mathbf{Y}\,]\big)$ \Comment{embed full sequence, $P+P_f$ patches}
\State $\mathbf{Z}^{\mathrm{teacher}}\gets \mathrm{EncoderMFVI}(\mathbf{Z}^{\mathrm{full}};\,K_{\mathrm{enc}})$ \Comment{causal MFVI over full seq; future leaks to no history position}
\Statex
\State \textit{\# Student path (same causal operator, history-only input)}
\State $\mathbf{Z}^{\mathrm{stu}}\gets \mathrm{EncoderMFVI}\!\big(\phi_u(\mathbf{X}_{\mathrm{hist}});\,K_{\mathrm{enc}}\big)$ \Comment{$P$ history latents; $\mathbf{Z}^{\mathrm{stu}}[{:}P]\!\equiv\!\mathbf{Z}^{\mathrm{teacher}}[{:}P]$}
\State initialise prediction list $\hat{\mathbf{Y}}\gets []$
\For{$k = 1,\dots, P_f$}
  \State $z_{\mathrm{prev}}\gets \mathbf{Z}^{\mathrm{stu}}[:,:,P+k-1,:]$ \Comment{most recent cached latent}
  \State $\hat{\mathbf{x}}_{k}\gets \mathrm{MLP}_{\mathrm{pred}}(z_{\mathrm{prev}})$;\quad append to $\hat{\mathbf{Y}}$ \Comment{emit next patch}
  \State $u_{\mathrm{patch}}\gets \phi_u(\hat{\mathbf{x}}_{k})$ \Comment{re-encoded prediction (observation evidence)}
  \State $u_{\mathrm{trans}}\gets p_\psi(z_{\mathrm{prev}})$ \Comment{latent transition prior}
  \State $z_{\mathrm{new}}\gets \mathrm{SingleStepMFVI}\!\big(u_{\mathrm{patch}}+u_{\mathrm{trans}},\;\mathbf{Z}^{\mathrm{stu}},\;\phi_b;\,K_{\mathrm{dec}}\big)$
  \Statex \hspace{1.1em}\Comment{fuses patch+trans unary with temporal cache, cross-channel msgs, global topic}
  \State $z_{P+k}\gets \alpha\cdot z_{\mathrm{prev}} + (1-\alpha)\cdot z_{\mathrm{new}}$ \Comment{damped update}
  \State append $z_{P+k}$ to $\mathbf{Z}^{\mathrm{stu}}$ \Comment{self-output is fed back into next step}
\EndFor
\Statex
\State $\mathcal{L}_{\mathrm{pred}}\gets \mathrm{MSE}(\hat{\mathbf{Y}},\,\mathbf{Y})$
\State $\mathcal{L}_{\mathrm{distill}}\gets \mathrm{SmoothL1}\!\big(\mathbf{Z}^{\mathrm{stu}}[:,:,P{:},:],\; \mathbf{Z}^{\mathrm{teacher}}[:,:,P{:},:]\big)$ \Comment{future only; history latents already coincide by causality}
\State \Return $\mathcal{L}\gets \mathcal{L}_{\mathrm{pred}} + \lambda_{\mathrm{latent}}\cdot \mathcal{L}_{\mathrm{distill}}$
\end{algorithmic}
\end{algorithm}

\end{document}